\documentclass[10pt,twocolumn,letterpaper]{article}

\usepackage[pagenumbers]{cvpr}              

\definecolor{ourgreen}{RGB}{46, 204, 113}
\definecolor{ourgreenborder}{RGB}{39, 174, 96}
\definecolor{ourblue}{RGB}{52, 152, 219}
\definecolor{ourblueborder}{RGB}{41, 128, 185}
\definecolor{ourorange}{RGB}{230, 126, 34}
\definecolor{ourorangeborder}{RGB}{211, 84, 0}
\definecolor{ourred}{RGB}{231, 76, 60}
\definecolor{ourredborder}{RGB}{192, 57, 43}
\definecolor{ouryellow}{RGB}{241, 196, 15}
\definecolor{ouryellowborder}{RGB}{243, 156, 18}
\definecolor{ourpurple}{RGB}{155, 89, 182}
\definecolor{ourpurpleborder}{RGB}{142, 68, 173}
\definecolor{ourturquoise}{RGB}{26, 188, 156}
\definecolor{ourturquoiseborder}{RGB}{22, 160, 133}
\definecolor{ourturquoise}{RGB}{26, 188, 156}
\definecolor{ourturquoiseborder}{RGB}{22, 160, 133}
\definecolor{ourwhite}{RGB}{236, 240, 241}
\definecolor{ourwhiteborder}{RGB}{189, 195, 199}
\definecolor{ourgray}{RGB}{149, 165, 166}
\definecolor{ourgrayborder}{RGB}{127, 140, 141}
\definecolor{sggreen}{HTML}{6ACC64}

\newcommand{\modelname}{\textsc{TR-DiT-2.5B}\xspace}
\definecolor{lightgray}{gray}{0.6}
\definecolor{lightgrayrow}{gray}{0.9}

\usepackage[most]{tcolorbox}

\tcbset{
  on line,
  boxsep=1pt,
  left=2pt,
  right=2pt,
  top=0.5pt,
  bottom=0.5pt,
  arc=3pt,
  boxrule=0pt,
  coltext=black,
}

\newtcbox{\maskingbox}{colback=orange!30!white}
\newtcbox{\routingbox}{colback=blue!20!white}

\definecolor{cvprblue}{rgb}{0.21,0.49,0.74}
\usepackage[utf8]{inputenc}
\usepackage[T1]{fontenc}
\usepackage[table,dvipsnames,x11names]{xcolor}
\usepackage[pagebackref,breaklinks,colorlinks,allcolors=cvprblue]{hyperref}
\usepackage{url}
\usepackage{booktabs}
\usepackage{amsfonts,amssymb,amsmath,mathtools}
\usepackage{nicefrac}
\usepackage{microtype}
\usepackage{graphicx}
\usepackage{subcaption}
\usepackage[capitalize]{cleveref}
\usepackage{adjustbox}
\usepackage{pifont}
\usepackage{dblfloatfix}
\usepackage{caption}
\usepackage{cuted}
\usepackage{capt-of}
\usepackage{wrapfig}
\usepackage{makecell}
\usepackage{soul} 
\usepackage[normalem]{ulem}
\usepackage{tikz}
\usetikzlibrary{calc}

\usepackage{tikz}
\usetikzlibrary{positioning,calc}
\usetikzlibrary{calc,fit,positioning,arrows.meta,decorations.pathmorphing, tikzmark, matrix}
\newcommand{\mk}[2]{\tikzmarknode{#1}{#2}}
\newcommand{\coloruline}[2]{%
  \tikz[baseline=(X.base)]{
    \node[inner sep=0pt, outer sep=0pt] (X) {#2};
    \draw[#1, line width=0.8pt]
      ([yshift=-1.2pt]X.south west) -- ([yshift=-1.2pt]X.south east);
  }%
}

\def\paperID{REMOVED} 
\def\confName{CVPR}
\def\confYear{2026}

\title{Guiding Token-Sparse Diffusion Models}

\author{
    Felix Krause\hspace{0.8em}Stefan Andreas Baumann\hspace{0.8em}Johannes Schusterbauer \\
    \hspace{0.8em}Olga Grebenkova\hspace{0.8em}Ming Gui\hspace{0.8em}Vincent Tao Hu\hspace{0.8em}Björn Ommer\\[1ex]
    CompVis @ LMU Munich, \quad Munich Center for Machine Learning (MCML)\\
}

\begin{document}
\maketitle

\begin{strip}
\centering
\includegraphics[width=\textwidth]{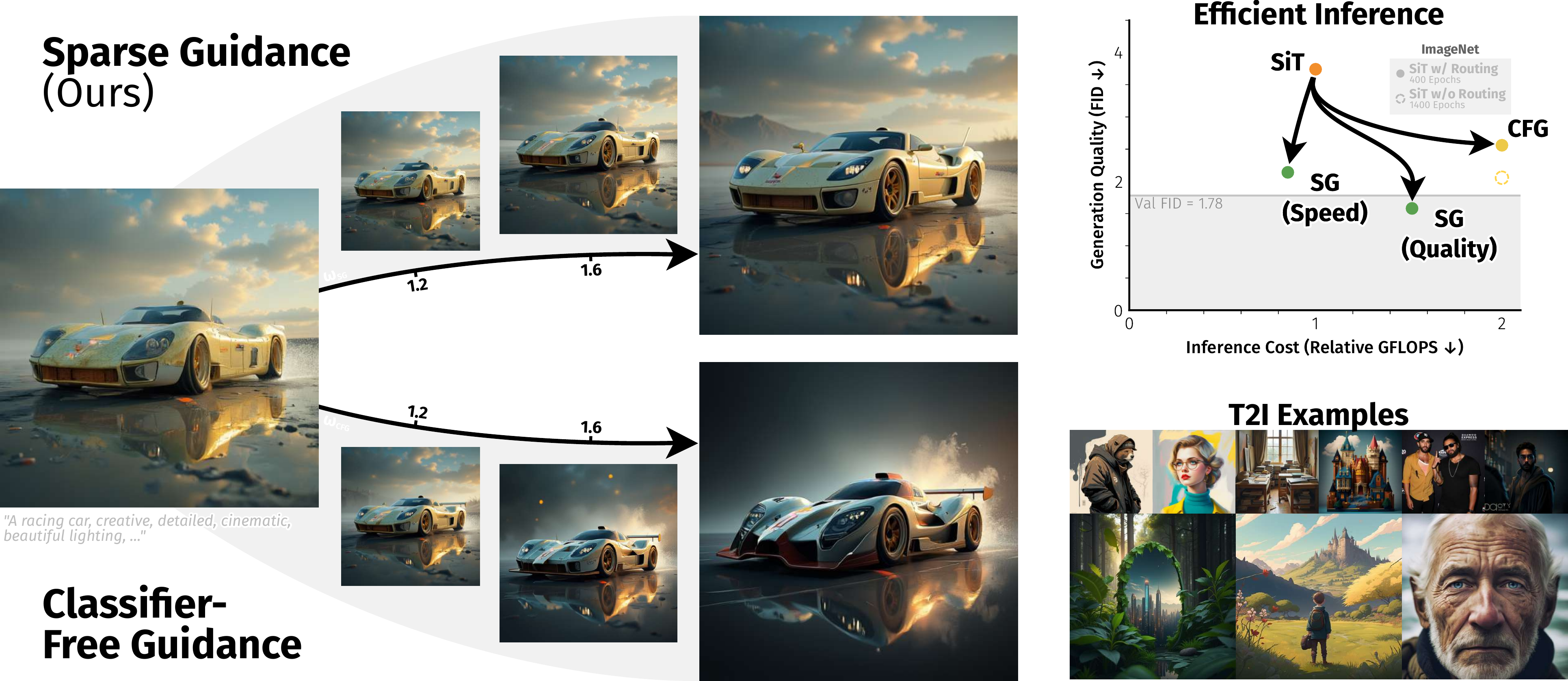}
\captionof{figure}{\textbf{Sparse Guidance provides effective, efficient, structure-preserving guidance for sparsely trained diffusion models.} (Left) Unlike Classifier-free Guidance, SG stays closer to the conditional prediction, yielding higher-variance, non-collapsed samples. (Right, top) On ImageNet-256, SG (Quality) attains an FID of \emph{1.58} without any previously required dense finetuning while also increasing throughput, and SG (Speed) matches the baseline quality at substantially lower inference cost. (Right, bottom) Applied to our 2.5B text-to-image model, Sparse Guidance raises its HPSv3 \cite{ma2025hpsv3} performance enough to surpass a range of larger models, which it could not achieve without SG.
}
\label{fig:teaser}
\vspace{-1em}
\end{strip}

\begin{abstract}
    Diffusion models deliver high quality in image synthesis but remain expensive during training and inference. Recent works have leveraged the inherent redundancy in visual content to make training more affordable by training only on a subset of visual information. While these methods were successful in providing cheaper and more effective training, sparsely trained diffusion models struggle in inference. This is due to their lacking response to Classifier-free Guidance~(CFG) leading to underwhelming performance during inference. To overcome this, we propose Sparse Guidance~(SG). Instead of using conditional dropout as a signal to guide diffusion models, SG uses token-level sparsity. As a result, SG preserves the high-variance of the conditional prediction better, achieving good quality and high variance outputs. Leveraging token-level sparsity at inference, SG improves fidelity at lower compute, achieving 1.58 FID on the commonly used ImageNet-256 benchmark with 25\% fewer FLOPs, and yields up to 58\% FLOP savings at matched baseline quality. To demonstrate the effectiveness of Sparse Guidance, we train a 2.5B text-to-image diffusion model using training time sparsity and leverage SG during inference. SG achieves improvements in composition and human preference score while increasing throughput at the same time.\newline 
    Project Page: \url{https://compvis.github.io/sparse-guidance}
\end{abstract}
\begin{figure*}
  \centering
  \includegraphics[width=1.0\linewidth]{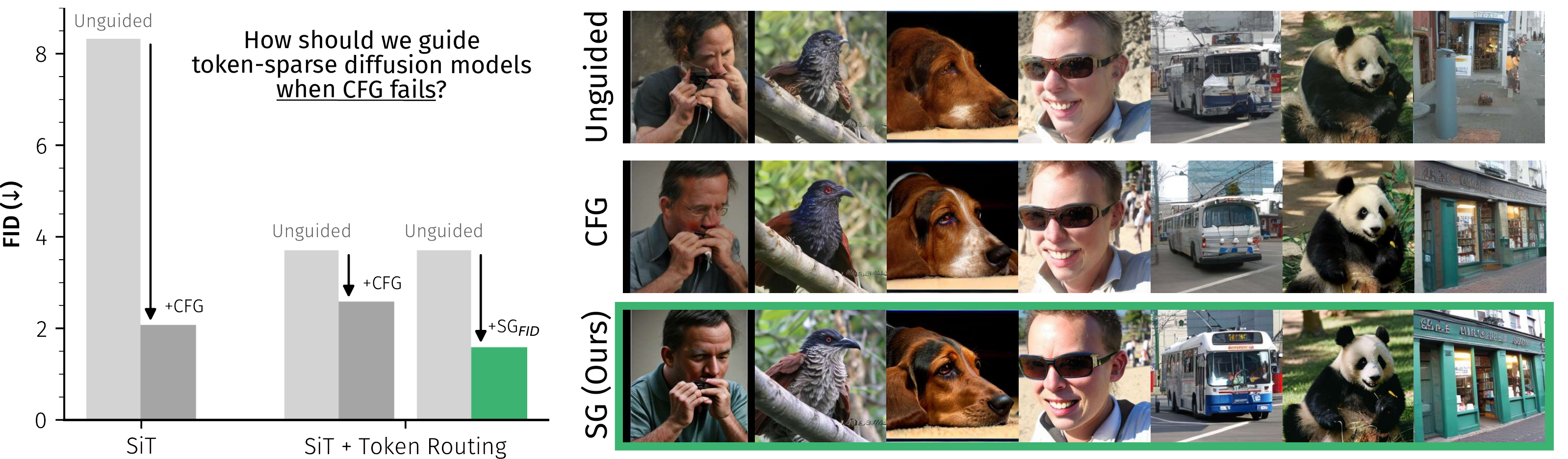}
    \caption{\textbf{Classifier-free Guidance (CFG) provides limited benefits for token-sparse diffusion models.} 
    While token-sparse training produces stronger conditional diffusion models than standard dense training, their practical impact has been constrained by poor compatibility with CFG, which limits inference quality and slows adoption in practice. \textbf{Sparse Guidance (SG) overcomes this limitation, restoring strong guidance gains for token-sparse models and enabling them to match or surpass the image quality of their dense baselines.}}
    
  \label{fig:motivation_figure}
\end{figure*}

\vspace{-4mm}
\section{Introduction}
In recent years, models developed by the machine learning community and industry have grown dramatically in size, thereby demanding massive computational resources~\citep{flux2024,videoworldsimulators2024,veo2,hurst2024gpt4o}. Diffusion models~\citep{sohl2015deep,ho2020denoising,lipman2022flow} have become a frequently used standard across modalities such as images \cite{rombach2022high_latentdiffusion_ldm,esser2024scalingrectifiedflowtransformers,flux2024} and video \cite{videoworldsimulators2024,veo2, blattmann2023stable}, despite being among the most compute-intensive approaches. Furthermore, Classifier-free Guidance (CFG) is commonly used for high generation quality. During CFG, an unconditional and a conditional prediction are combined, which typically doubles the inference costs of already very expensive diffusion models \cite{ho2021classifier}.

For the training of these models, methods like training-time sparsity~\citep{zheng2023fast_maskdit, krause2025tread,Gao_2023_ICCV} have shown improvements in efficiency as well as performance. These methods exploit the underlying redundancy of visual data and train a diffusion model only on a subset of available information at any given time. \emph{Masking} replaces the discarded information with learnable parameters while \emph{routing} aims to first withdraw and later reintroduce information. The reason the community has not adopted these approaches fully is a breakdown of inference capabilities: models trained with such training-time sparsity show unreliable and often weak performance during generation due to their unresponsiveness to CFG~\citep{zheng2023fast_maskdit, krause2025tread, zhu2024sddit}.

We propose \emph{Sparse Guidance} (SG) as a direct remedy to the issue of costly inference and the practical usability of sparsely trained diffusion models at the same time. SG steers the generation process by leveraging a \emph{capacity gap} induced by inference-time sparsity (i.e., a controlled difference between two predictions created by two distinct token-level sparsity rates). Unlike previous approaches \cite{zheng2023fast_maskdit, krause2025tread, sehwag2024stretching}, SG requires no additional finetuning to recover the model’s capabilities under CFG while providing \textbf{higher quality} with \textbf{better throughput} as Sparse Guidance embraces the train-test gap of sparse training approaches instead of avoiding it. We validate SG on the commonly used ImageNet-256 benchmark, where SG achieves an FID of 1.58. Furthermore, we show predictable behavior and a smooth quality–throughput trade-off, where increasing inference-time sparsity reduces the number of processed tokens and lowers computational cost. Then we demonstrate that SG holds up at scale: we train a 2.5B text-to-image Diffusion Transformer using token routing~\citep{krause2025tread} and, applying SG, find reliable improvements in image quality measured by human preference, alongside reduced FLOPs and increased inference throughput.

\noindent Our main contributions can be summarized as:
\begin{itemize}
    \item We introduce \emph{Sparse Guidance} (SG), a finetune-free, post-hoc scheduling mechanism for sparsely trained diffusion models. SG computes two predictions and applies token-level sparsity to them and then utilizes their capacity gap to steer the generation towards higher quality. As tokens are removed from the computational branch, the cost for inference shrinks naturally. 
    \item Sparse Guidance delivers strong results without additional finetuning. SG achieves \textbf{FID 1.58} with \textbf{25\%} fewer FLOPs, and up to \textbf{58\%} savings at comparable quality to a dense SiT on the commonly used ImageNet-256 benchmark.
    \item To demonstrate the viability of this pipeline, we train a large scale text-to-image 2.5B Diffusion Transformer using token routing. We apply our proposed Sparse Guidance method which improves image quality measured by human preference score and naturally increases throughput during inference significantly by reducing the amount of processed information. 
\end{itemize}
\section{Related Work}

\paragraph{Diffusion and Flow Matching Models.}
Score-based diffusion models, such as DDPM~\citep{ho2020denoising} and its improved variants~\citep{song2020improved, nichol2021improved, song2020score, song2020denoising_ddim}, as well as Latent Diffusion Models~\citep[LDM,][]{rombach2022high_latentdiffusion_ldm}, have become the cornerstone of high-fidelity synthesis across images~\citep{ramesh2022hierarchical, schusterbauer2024boosting}, video~\citep{ho2022video, bar2024lumiere} and audio~\citep{liu2023audioldm, huang2023make, nistal2024diff}.
Complementarily, flow-matching methods~\citep{lipman2022flow, rectifiedflow_iclr23, albergo2023stochastic, ma2024sit} recast generation as learning a continuous vector field within an interpolant framework that unifies flow and diffusion, enabling efficient ODE-based sampling.
Early diffusion frameworks relied on U-Net backbones~\citep{unet}, but recent work has shifted toward token-based transformers like DiT~\citep{dit_peebles2022scalable}, which offer scalability at the cost of quadratic complexity in the number of tokens~\citep{zheng2023fast_maskdit}. 
To mitigate this, caching schemes accelerate inference in both U-Nets~\citep{ma2023deepcache} and DiTs~\citep{ma2024learningtocacheacceleratingdiffusiontransformer}, yet still process every token at each layer.
In contrast, we utilize a test-time token-sparsity which allows us to reduce the number of processed tokens per layer.

\paragraph{Diffusion Guidance.}
Guidance has become a standard tool for improving the fidelity of diffusion model outputs. 
An auxiliary model or signal steers the generative process~\citep{dhariwal2021diffusion}. 
Currently, the most dominant approach is classifier-free guidance (CFG)~\citep{ho2021classifier}, which combines the conditional and unconditional score to improve sample fidelity at the cost of diversity. 
Recent advances such as Autoguidance (AG)~\citep{karras2024guiding} use a smaller and less trained model to replace the previously used unconditional branch to achieve good guidance. 
\citet{sadat2024no} apply perturbations to the timestep embeddings, causing intentional misalignment in noise removal to guide the generation process. 
\citet{kaiser2024unreasonable} restrict the receptive field in convolution-based backbones for guidance. 
Beyond these classifier- and branch-based methods, attention-based schemes such as self-attention guidance \cite{hong2023improving} and perturbed-attention guidance \cite{ahn2024self} steer sampling by manipulating internal attention patterns. 
In contrast to previous methods, we propose to apply train-time sparsity augmentations to inference by using two token-sparsity rates (number of concurrently processed tokens) to create a capacity gap which we effectively use to steer the sampling process towards higher quality. 

\paragraph{Token Sparsity.}
In parallel, efficiency-focused research has enabled models like the Transformer~\cite{vaswani2017attention} to skip processing of less important tokens.
Token masking has shown that the entire token set is not required for a diffusion model to approximate the data distribution~\citep{zheng2023fast_maskdit, zhu2024sddit, Gao_2023_ICCV}.
The advantage in these methods is that training throughput is increased significantly, which reduces costs.
As an alternative to masking, token routing reintroduces tokens instead of replacing them with learnable embeddings~\citep{krause2025tread}.
In the domain of diffusion models, such routing can preserve token information, providing better convergence speed while retaining the efficiency of similar masking methods.
Relatedly, Mixture-of-Depths~\citep{raposo2024mixture} employs a fixed top-$k$ token selection per layer, which allows only $k$ tokens to be processed by each layer, reducing computational cost.
Beyond train-time masking and routing, test-time token merging and pruning in diffusion transformers reduce compute by compressing or dropping tokens while preserving visual quality.
Furthermore, feature-caching approaches such as DeepCache and Learning-to-Cache accelerate diffusion U-Nets and transformers by reusing intermediate activations across timesteps or layers~\citep{ma2023deepcache, ma2024learningtocacheacceleratingdiffusiontransformer}.
Our method builds on train-time sparsity but introduces it to inference leveraging it as a guidance signal to improve visual quality.

\section{Method}
\label{sec:method}

\subsection{Preliminaries}
\label{subsec:prelim}

\begin{figure}[t]
    \centering
    \includegraphics[width=\linewidth]{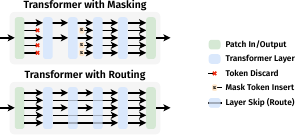}
    \caption{\textbf{Masking and Routing as two types of token-level sparsity.} Masking replaces tokens with learnable mask token \cite{zheng2023fast_maskdit} while routing preserves information by reintroducing tokens \cite{krause2025tread}.}
    \label{fig:masking_routing}
    \vspace{-2mm}
\end{figure}

\paragraph{Flow Matching.}
Flow Matching (FM) formulates generation as learning a continuous-time vector field that deterministically transports a simple prior distribution to the data distribution \cite{lipman2022flow, albergo2023stochastic,rectifiedflow_iclr23}. Concretely, let $z \sim \mathcal{N}(0,I)$ denote a latent sample from the prior and $x \sim p_{\text{data}}$ a corresponding data sample. We adopt the widely used standard straight (Gaussian) interpolation path \cite{lipman2022flow}
\begin{equation}
x_t \;=\; (1-t)\,z \;+\; t\,x,\qquad t\in[0,1],
\label{eq:path}
\end{equation}
whose oracle velocity is constant along the path,
\begin{equation}
v^\star(x_t,t) \;=\; \frac{d x_t}{dt} \;=\; x - z.
\label{eq:oracle}
\end{equation}
A flow-matching model $v_\theta$ predicts $v^\star$, and sampling integrates the ODE $\frac{d x_t}{dt}=v_\theta(x_t,t)$ from $t=0$ to $t=1$ \cite{lipman2022flow}.

\paragraph{Classifier-free Guidance}
High-fidelity sampling often employs \emph{Classifier-free Guidance} (CFG) to steer the conditional prediction away from a weaker (unconditional) branch. For brevity, we write \(v_\theta(x_t,t,c)\) as \(v_\theta(c)\) and retain only guidance-relevant terms. Given conditioning \(c\) and guidance scale \(\omega \geq 1\), Classifier-free Guidance \cite{ho2021classifier} is defined as:
\begin{equation}
v_{\theta}^{\text{CFG}}(c, \omega) \;=\; \omega\,v_\theta(c)\;+\;(1-\omega)\,v_\theta(\varnothing).
\label{eq:cfg}
\end{equation}
CFG doubles per-step compute for dense models. Our goal is to retain its benefits while \emph{reducing} the compute increase under sparsity.

\paragraph{Token Sparsity.}
Let \(D_{\theta}\) denote the denoiser network, composed of \(B\) sequential layers \(L_0, \ldots, L_{B-1}\). Token sparsity reduces training cost by avoiding computation on the full set of tokens in every layer:  \emph{Masking} drops a fixed fraction \(\gamma\) of tokens and optionally replaces them with learnable embeddings, never re-inserting the original activations. We then define masking as follows:
\begin{equation}
    D_{\theta}^{\textbf{m}} = L_{B-1} \circ \cdots \circ \left\{
    \begin{array}{ll}
        \!\!\operatorname{mask}, & \!\!\tau_k \in \mathcal{T}_{\textbf{m}}\!\!\\[1mm]
        \!\! L_k \circ \cdots \circ L_0, & \!\!\text{otherwise}\!\!
    \end{array}
    \right\}\!,
\end{equation}
where \(\operatorname{mask}(\tau_k) = e_{\text{mask}}\) replaces token \(\tau_k\) with a fixed or learnable embedding that carries no instance-specific information, permanently removing the original activation from the forward path.

\emph{Routing} selects a subset of tokens to process and re-inserts them later, keeping all tokens within the computational graph. This is then defined as:
\begin{equation}
    D_{\theta}^{\textbf{r}_{i\rightarrow j}} = L_{B-1} \circ \cdots \circ \left\{
    \begin{array}{ll}
        \!\!\!\operatorname{id}, & \!\!\tau_k \in \mathcal{T}_{\textbf{r}_{i \rightarrow j}}\!\!\!\!\\[1mm]
        \!\!\! L_j \circ \cdots \circ L_i, & \!\!\text{otherwise}\!\!\!\!
    \end{array}
    \right\} \circ \cdots \circ L_0,
\end{equation}
where \(\operatorname{id}\) denotes the identity mapping applied to routed tokens, ensuring they bypass intermediate layers while preserving their information for later re-insertion. \Cref{fig:masking_routing} demonstrates this visually.

\subsection{Sparse Guidance (SG)}
\label{subsec:sg}

\paragraph{Using Training Augmentation as a Guidance Signal.}
Token-level sparsity has proven effective for accelerating \emph{training}~\cite{Gao_2023_ICCV,krause2025tread,zhu2024sddit}. However, at \emph{inference} time, models employing standard classifier-free guidance (CFG) frequently exhibit decreased response to the guidance signal or degraded fidelity unless subjected to dense finetuning (see \Cref{fig:motivation_figure}). We revisit sparsity not as a training-only device but as a \emph{test-time control signal}. Formally, let $\gamma\in[0,1)$ denote a sparsity rate that either masks tokens (replacement by a fixed/learnable embedding) or routes tokens (bypassing selected layers with identity and later reinsertion). 

\begin{figure}[t]
    \centering
    \includegraphics[width=1.0\linewidth]{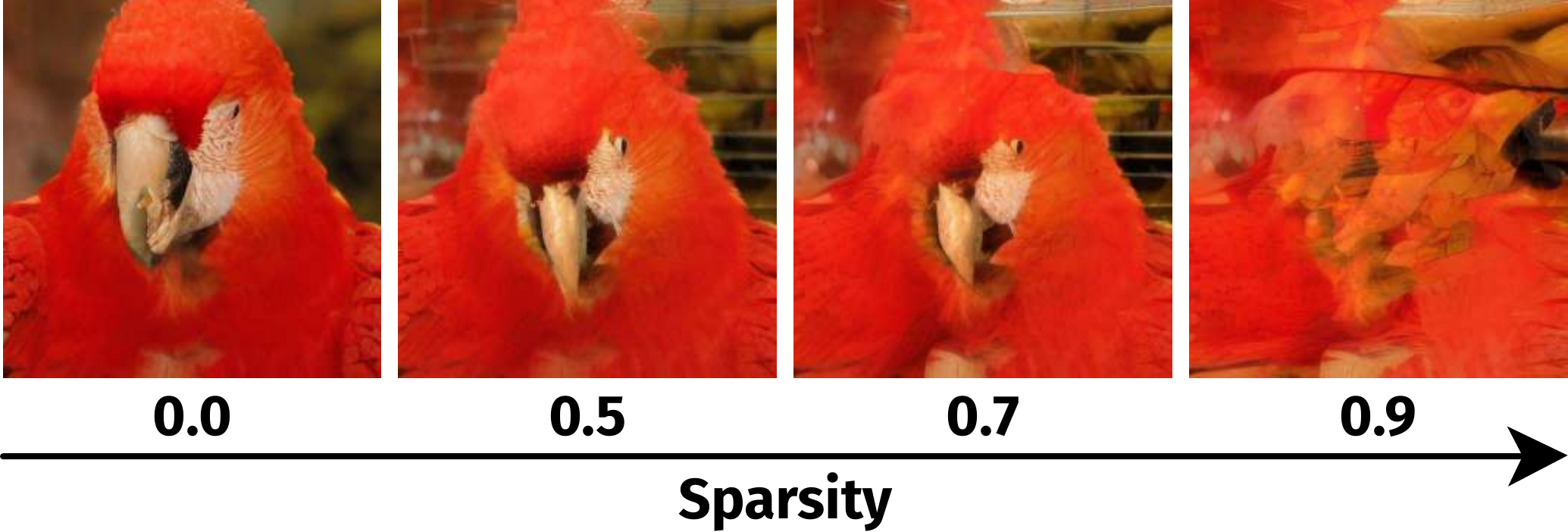}
    \caption{\textbf{Without Sparse Guidance, image quality and composition worsens consistently with increased token-sparsity ratios.}}
    \label{fig:unguided_sparsity}
    \vspace{-2mm}
\end{figure}

\paragraph{Controlling Capacity with Sparsity.}
Naively adapting a token-level sparsity $\gamma>0$ during inference ($\omega=1.0$) leads to deteriorated outputs \Cref{fig:unguided_sparsity}. As $\gamma$ increases, the model’s effective capacity shrinks, limiting its ability to realize the learned distribution and producing visually disturbing artifacts. To overcome this, we utilize the capacity-controlling sparsity knob $\gamma$ during inference only in a guided setting. Guidance is most effective when a high-variance predictor pushes a lower-variance one toward outputs with even less variance (e.g., a specific conditioning)~\cite{ho2021classifier,karras2024guiding,kynkäänniemi2024applyingguidancelimitedinterval}. We find that token-level sparsity provides a direct knob for realizing this: increasing $\gamma$ lowers effective capacity and \emph{softens} the conditional distribution produced by $D_\theta(x_t,t,c;\gamma)$, while decreasing $\gamma$ yields a sharper, higher-capacity predictor. We propose instantiating guidance by using a high-$\gamma$ (weak) branch to steer a low-$\gamma$ (strong) branch during sampling. The resulting capacity gap provides the guidance signal. In this view, $\gamma$ is a single, continuous hyperparameter over distributional sharpness, turning train-time sparsity into a test-time \emph{guidance primitive}.

\paragraph{Guidance Formulation.}
We evaluate the network $D_\theta$ under two test-time sparsity levels using the notation $D_\theta(x_t,t,c;\gamma)$ to indicate token sparsity $\gamma$. Further, we will define the two branches that are needed for a guided prediction as $D_\theta^{\text{strong}}$ and $D_\theta^{\text{weak}}$, no matter what $\gamma_\text{strong}$ or $\gamma_\text{weak}$ is applied respectively.
\begin{equation}
\label{eq:sg-def-capacities}
\begin{aligned}
D_\theta^{\text{strong}}(c) &:= D_\theta(x_t,t,c;\gamma_{\text{strong}}),\\
D_\theta^{\text{weak}}(c) &:= D_\theta(x_t,t,c;\gamma_{\text{weak}}),\\
&\quad 0 \le \gamma_{\text{strong}} < \gamma_{\text{weak}} < 1.
\end{aligned}
\end{equation}
In contrast to CFG, both predictions are conditional. Consequently, the guidance signal is provided solely by the capacity gap induced by the difference in sparsity $\gamma_{\mathrm{s}} \not=\gamma_{\text{weak}}$.
Then we utilize the guidance formulation,
\begin{equation}
\label{eq:sg}
\begin{aligned}
D_\theta^{\mathrm{SG}}\!\left(c,\gamma_{\text{strong}},\gamma_{\text{weak}},\omega\right)
&= \omega\, D_\theta^{\text{strong}}(c) \\
&\quad + (1-\omega)\, D_\theta^{\text{weak}}(c)
\end{aligned}
\end{equation}
which uses the low-capacity, weak prediction $D_\theta^{\text{weak}}(c)$ to steer the high-capacity, strong prediction in the direction of
$D_\theta^{\text{strong}}(c) - D_\theta^{\text{weak}}(c)$ with magnitude $\omega$.

\begin{figure*}[h!]
  \centering
  \begin{subfigure}[t]{0.475\linewidth}\vspace{0pt}
    \centering
    \label{fig:fid_vs_steps}
    \adjustbox{max width=\linewidth}{
      \includegraphics{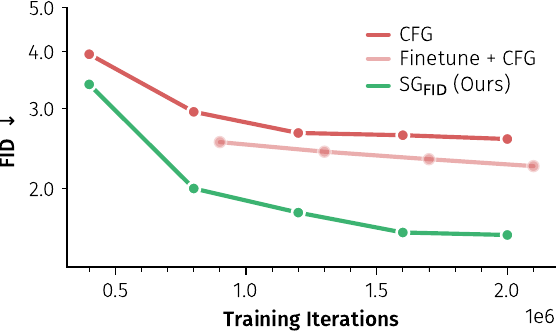}
    }
  \end{subfigure}\hfill
  \begin{subfigure}[t]{0.505\linewidth}\vspace{0pt}
    \centering
    \label{fig:training_progress}
    \includegraphics[width=\linewidth]{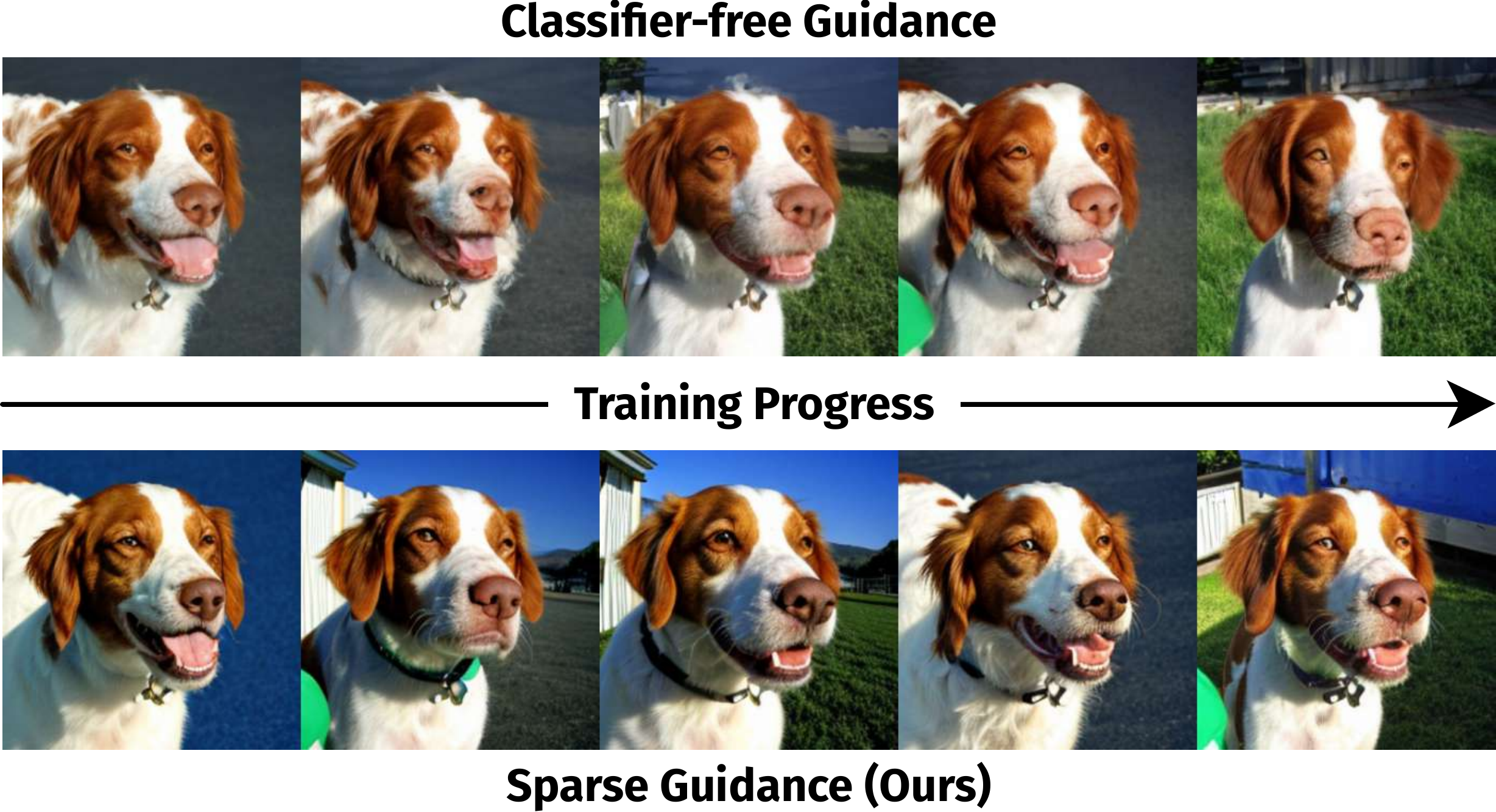}
  \end{subfigure}
  \vspace{-3mm}
  \caption{\textbf{Sparse Guidance improves both convergence and training-time sample quality for sparsely trained diffusion models.} \textbf{Left:} FID over training iterations comparing CFG, CFG with dense finetuning, and Sparse Guidance (SG), where SG achieves the lowest FID. We use the best $\omega$ for each method. \textbf{Right:} Training-time sample progress using SG, showing that sparsely trained models already produce high-fidelity samples without an additional dense finetuning stage, enabling direct visual evaluation during training.}
  \label{fig:sg_combined}
  \vspace{-3mm}
\end{figure*}

As SG makes no assumptions about the provided conditioning, it can be combined naturally with other existing guidance techniques. Applying the zero-condition $\varnothing$ to our weak branch leads to the combination of Classifier-free Guidance and Sparse Guidance (CFG + SG):
\begin{equation}
\label{eq:sgcfg}
\begin{aligned}
D_\theta^{\mathrm{CFG+SG}}\!\left(c,\gamma_{\text{strong}},\gamma_{\text{weak}},\omega\right)
&= \omega\, D_\theta^{\text{strong}}(c) \\
&\quad + (1-\omega)\, D_\theta^{\text{weak}}(\varnothing).
\end{aligned}
\end{equation}
Similarly, SG can be combined with other guidance signals, like AutoGuidance \cite{karras2024guiding} where $D_\theta^{\text{weak}}$ is represented by a smaller model or undertrained checkpoint. 

At test time, token subsets are sampled from binary masks $m \in \{0,1\}^T$ with $m_k \sim \mathrm{Bernoulli}(1-\gamma)$ for $\gamma \in \{\gamma_{\text{strong}}, \gamma_{\text{weak}}\}$.

\paragraph{Hyperparameter Usage.}
Prior works applying sparsity during training often come with a variety of additional hyperparameters with their respective sparsity (or masking) rate being one of them \cite{Gao_2023_ICCV, zheng2023fast_maskdit, krause2025tread}. Furthermore, several other guidance methods require affected layers to be handpicked for effective guidance \cite{ahn2024self, hyung2025spatiotemporal} while Sparse Guidance copies the train-time settings and applies them during inference leaving only $\gamma$ as additional hyperparameter.

\section{Experiments}
We test our proposed Sparse Guidance method to leverage sparsely trained diffusion models during inference. To that end, we evaluate on class-conditional ImageNet-256 generation across model scales and compare to relevant guidance based baselines. Further, we provide evidence that Sparse Guidance and thereby indirectly sparse training methods as well, scale to billion parameter sized text-to-image models.

\begin{figure*}[t]
    \centering
\includegraphics[width=\linewidth]{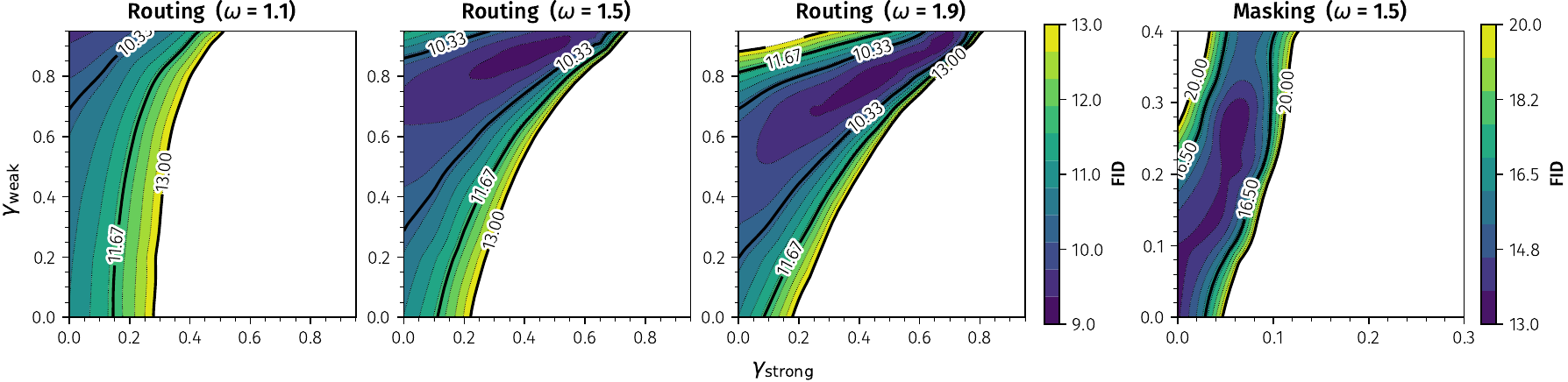}
    \caption{\textbf{Our method achieves lower FID robustly across different $\omega$} by adaptation of $\gamma_{\text{strong}}$ and  $\gamma_{\text{weak}}$. We demonstrate how SG allows for fine grained control over the capacity gap between the $D_\theta^{\text{strong}}$ and $D_\theta^{\text{weak}}$ that drives guidance. Notably, the area of viable settings is broad and shifts under increasing $\omega$ towards higher $\gamma_{\text{strong}}$ and  $\gamma_{\text{weak}}$.}
    \label{fig:maskdit_tread_combined}
\end{figure*}

\subsection{Experimental Setup}
\paragraph{ImageNet.} Our experimental setup follows standard evaluation protocols, evaluating models in the class-conditional latent ImageNet-$256^2$ setting that the various methods~\citep[][]{zheng2023fast_maskdit,krause2025tread,dit_peebles2022scalable} were developed for. To enable fair comparisons, we reproduce both a masking~\citep[MaskDiT,][]{zheng2023fast_maskdit} and routing~\citep[TREAD,][]{krause2025tread} model with the settings proposed in the respective works. We train using AdamW~\citep{loshchilov2017decoupled_adamw} at a learning rate of $1\times 10^{-4}$ at a batch size of 256 with default betas $(\beta_{1}, \beta_{2}) = (0.9, 0.999)$. We train both models as SiT-XL/2~\citep{ma2024sit,dit_peebles2022scalable} models in the latent space of the Stable Diffusion~\citep{rombach2022high_latentdiffusion_ldm} VAE. During inference, we sample using a simple euler sampler with 40 steps, unless noted otherwise. We evaluate samples using the standard established evaluation protocol, primarily relying on the Fr\'echet Inception Distance~\citep[FID,][]{heusel2017gans_fid} for evaluation of generated sample quality. We use the standard implementation from ADM~\citep{dhariwal2021diffusion} and, unless noted otherwise, compute FID based on 50k random samples. In addition to FID, we also report sFID~\citep{nash2021generating}, Inception Score~\citep[IS,][]{salimans2016improved_is_inceptionscore}, and Precision and Recall~\citep{kynkaanniemi2019improved} for our main results.
We report further implementation details and comprehensive descriptions and details for all shown results achieved with Sparse Guidance in the Appendix (\Cref{sec:implementation}).

\paragraph{Scaling Up to Text-to-Image (T2I).} To test if Sparse Guidance works beyond ImageNet with small to medium-sized models, we train a 2.5B text-to-image diffusion transformer. We utilize the internVL3-2b \cite{zhu2025internvl3} model as text encoder and apply a prompt prefix and insert a two layer transformer network between the Vision Language Model (VLM) and the Cross-Attention of our DiT as proposed by \citet{ma2024exploring}. We use TREAD \cite{krause2025tread} as our training time sparsity and follow the proposed settings with a route from $L_2\rightarrow L_{30}$ in a 34 layer network and 50\% selection rate. We train our model on a recaptioned subset of COYO-700M \cite{kakaobrain2022coyo-700m} which sums up to 100M samples. We divide our training into two stages. In the first, we train on all 100M samples while in the second one, we filter our data according to aesthetics score and add synthetic data from JourneyDB \cite{pan2023journeydb} and FLUX-6M \cite{fang2025flux}. During inference, we use a 512$\times$512 resolution with 50 euler sampling steps and apply \texttt{bfloat16}.

\subsection{Sparse Guidance on ImageNet}\label{sec:performance}
\paragraph{Sparse Approaches}
We apply our Sparse Guidance to models trained using state-of-the-art sparse training methods. As dropping tokens is a shared process among token-sparse methods, the differentiating factor becomes the replacement  of dropped tokens. We decide on masking \cite{zheng2023fast_maskdit} and routing \cite{krause2025tread} as they embody extreme cases (discard information vs. reuse). SG shows improved generative quality for both of these approaches which demonstrates broad applicability.
\begin{table}[h]
	\centering
	\adjustbox{max width=\linewidth}{
		\begin{tabular}{
			l@{\hskip 8pt}
			c@{\hskip 8pt}
			c@{\hskip 8pt}
			c@{\hskip 8pt}
			c@{\hskip 8pt}
			c@{\hskip 8pt}
			c@{\hskip 8pt}
			c@{\hskip 8pt}
		}
			\toprule
			\textbf{Guidance}
			& \textbf{Sparsity}
			& \textbf{\#Epoch}
			& \textbf{FID}$\downarrow$
			& \textbf{sFID}$\downarrow$
			& \textbf{IS}$\uparrow$
			& \textbf{Prec.}$\uparrow$
			& \textbf{Rec.}$\uparrow$ \\
			\midrule
			CFG       & masking & 160    & 5.82  & 13.00  & 227.8 & 0.80 & \textbf{0.45} \\
			\textbf{SG (Ours)} & masking & 160    & \textbf{5.73} & \textbf{11.99} & \textbf{249.0} & \textbf{0.83} & 0.42 \\
			\midrule
			CFG       & routing & 160    & 2.95  & \phantom{0}4.84   & \textbf{233.3} & \textbf{0.82} & 0.56 \\
			\textbf{SG (Ours)} & routing & 160    & \textbf{2.07}  & \phantom{0}\textbf{3.98}   & 223.4 & 0.80 & \textbf{0.58} \\ 
			\bottomrule
		\end{tabular}
	}
	\caption{\textbf{SG improves upon CFG for diffusion models trained with masking and routing as their train-time sparsity.}}
	\label{tab:masking_vs_routing}
\end{table}

\paragraph{Comparison against Guidance Methods.}
We evaluate \emph{Sparse Guidance} against a broad suite of guidance techniques for sparsely trained generators. Across all settings, both $\mathrm{SG}_{\mathrm{FID}}$ and $\mathrm{SG}_{\mathrm{FLOPS}}$ consistently outperform alternative guidance methods on the same sparsely pretrained backbone. Notably, $\mathrm{SG}_{\mathrm{FID}}$ achieves FID \textbf{$= 1.58$} at 400 epochs, yielding a further $0.99$ FID reduction over the next best competitor (CFG), by combining SG with AutoGuidance (AG). Beyond accuracy, SG reduces inference cost by enforcing sparsity at test time: $\mathrm{SG}_{\mathrm{FLOPS}}$ attains lower GFLOPs than the no-guidance baseline while surpassing the baseline’s quality with guidance. Under matched compute, SG also requires fewer operations than CFG, using \textbf{58\%} fewer GFLOPs (SG$_\text{FLOPS}$). Furthermore, we compare to Independent Condition Guidance (ICG) \cite{sadat2024no} which introduces a guidance method without requiring training interventions, unlike CFG. We find that, SG achieves better performance than ICG which underlines our claim that Sparse Guidance minimizes the train-test gap by introducing test-time sparsity. 
\begin{table}[b]
    \centering
    \label{tab:extensive_reduced}
    \adjustbox{max width=\linewidth}{\scalebox{.75}{
    \begin{tabular}{
        l
        c
        c
        r
        r
    }
    \toprule
    \textbf{Method}
    & \textbf{\#Epoch}
    & \textbf{FID}$\downarrow$
    & \textbf{GFLOPS}$\downarrow$
    & \textbf{$\Delta$GFLOPS}$\downarrow$ \\
    \midrule
    \rowcolor{gray!8}
    SiT-XL/2 + routing   & 400 & 4.89 & 114.42 & 0 (baseline)\\
    \hspace{1em}+CFG \cite{ho2021classifier}    & 400 & 2.57 & 228.84 & +114.42 \\
    \rowcolor{gray!8}
    \hspace{1em}+AG \cite{karras2024guiding}      & 400 & 2.95 & 228.84 & +114.42 \\
    \hspace{1em}+APG \cite{sadat2024eliminating} & 400 & 2.51  & 228.84  & +114.42                 \\
    \rowcolor{gray!8}
	\hspace{1em}+$S^2$ \cite{chen2025s}         & 400 & 3.12  & 228.84  & +114.42                 \\
    \hspace{1em}+ICG \cite{sadat2024no}     & 400 & 2.81 & 228.84 & +114.42 \\ 
    \rowcolor{gray!8}
    \hspace{1em}+$\text{SG}_{\text{FLOPS}}\,( \text{Ours} )$  & 400 & \underline{2.14} & \textbf{97.67} & \textbf{-16.75} \\
    \hspace{1em}+$\text{SG}_{\text{FID}}\,( \text{Ours} )$  & 400 & \textbf{1.58} & \underline{173.16} & \underline{+58.74} \\
    \bottomrule
    \end{tabular}}}
    \caption{\textbf{SG outperforms other guidance methods by significant margins} in FID and GFLOPS.}
\end{table}

\vspace{-5mm}
\paragraph{No Finetuning Requirements.} 
Prior works observe irregular behavior when applying classifier-free guidance (CFG) to sparsity-augmented diffusion models have reported that an additional \emph{dense} finetuning stage can partially restore CFG effectiveness \cite{zheng2023fast_maskdit, Gao_2023_ICCV, krause2025tread, sehwag2024stretching}. In \Cref{fig:sg_combined}, we show that even after an extensive dense finetuning phase, CFG still fails to match the performance of our proposed Sparse Guidance method. \Cref{fig:sg_combined} mirrors these metrics with visual results on the right. Consequently, this supports our central claim that SG is \emph{essential} to fully realize the generative capacity of sparsely trained diffusion models. 

\paragraph{State-of-the-Art Comparison.} 
Finally, we also compare with state-of-the-art diffusion models in \Cref{tab:compact}. Using our high-quality configuration $\text{SG}_{\text{FID}}$ , we achieve an FID of 1.58, outperforming a multitude of baselines while simultaneously offering a significant 24.6\% reduction in inference cost compared to a dense guided SiT baseline (173.16 vs 228.84 GFLOPS). Aside from FID, $\text{SG}_{\text{FID}}$ also provides larger recall \cite{kynkaanniemi2019improved}, indicating higher variance in sampled images.

\begin{table}[b]
    \centering
    \resizebox{\linewidth}{!}{%
      \begin{tabular}{lrccccc}
        \toprule
        \textbf{Method}        & \textbf{\#Epoch} & \textbf{FID}$\downarrow$ & \textbf{sFID}$\downarrow$ & \textbf{IS}$\uparrow$ & \textbf{Prec.}$\uparrow$ & \textbf{Rec.}$\uparrow$ \\
        \midrule

        DiT-XL/2  \cite{dit_peebles2022scalable}    
            & 1400   & 2.27            & 4.60             & 278.24       & \textbf{0.83}            & 0.57 \\

        \rowcolor{gray!8}
        SD-DiT-XL/2  \cite{zhu2024sddit} 
            & 480    & 3.23            & --               & --           & --              & --   \\

        FasterDiT-XL/2 \cite{yao2024fasterdit}
            & 400    & 2.03            & 4.63             & 264.00       & 0.81            & 0.60 \\

        \rowcolor{gray!8}
        MaskDiT-XL/2 \cite{zheng2023fast_maskdit} 
            & 1600   & 2.28            & 5.67             & 276.56       & 0.80            & 0.61 \\

        MDT-XL/2   \cite{Gao_2023_ICCV}   
            & 1300   & 1.79            & 4.57             & 283.01       & 0.81            & 0.61 \\

        \rowcolor{gray!8}
        SiT-XL/2 \cite{ma2024sit}      
            & 1400   & 2.06            & 4.50             & 270.30       & 0.82            & 0.59 \\

        SiT-XL/2 + REPA \cite{yu2024repa} 
            & 800    & 1.80            & 4.50             & \textbf{284.00}       & 0.81            & 0.61 \\
        \midrule
        \rowcolor{gray!8}
        SiT-XL/2 + routing \cite{krause2025tread}* 
            & 400    & 2.57            & 4.99             & 275.26       & 0.82            & 0.57 \\
        \hspace{10pt}+ $\text{SG}_{\text{FID}}\,( \text{Ours} )$ 
            & 400 & \textbf{1.58} & \textbf{4.45} & 249.70 & 0.80 & \textbf{0.63} \\

        \bottomrule
      \end{tabular}%
    }
    \caption{\textbf{SG achieves 1.58 FID on the ImageNet-256 benchmark.} * denotes our reproduced experiments.}
     \label{tab:compact}
     \vspace{-2mm}
\end{table}

\subsection{Effect of Sparsity}\label{sec:effect_of_sparsity}
At inference, we impose distinct sparsity rates on the two branches: $\gamma_{\text{strong}}$ on $D_\theta^{\text{strong}}$ and $\gamma_{\text{weak}}$ on $D_\theta^{\text{weak}}$. To study the behavior of these hyperparameters and their interaction with the guidance scale $\omega$, we evaluate the triplet $(\gamma_{\text{strong}}, \gamma_{\text{weak}}, \omega)$ across a range of combinations. For greater coverage of the configuration space, we report \mbox{FID@5k}, enabling a more exhaustive analysis than standard evaluation settings.

\begin{figure}[h!]
    \centering
    \includegraphics[width=\linewidth]{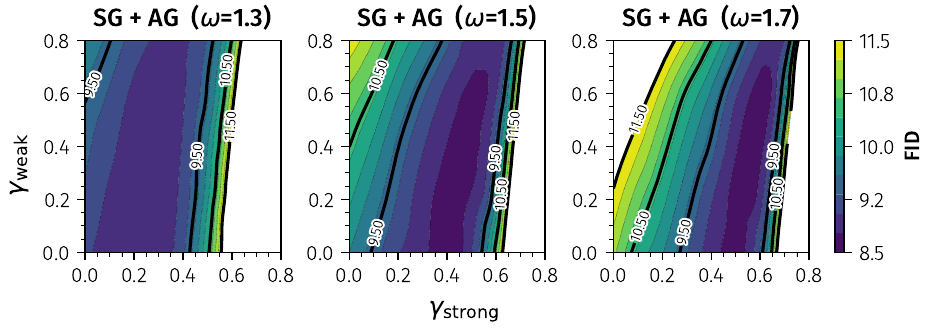}
    \caption{\textbf{Sparse Guidance combines well with other guidance signals} like AutoGuidance (AG) and demonstrates well behaved trade-off between $(\gamma_{\text{strong}}, \gamma_{\text{weak}}$ and $\omega)$ where larger $\omega$ allows for higher rates of sparsity and therefore also higher throughput.}
    \label{fig:routing_agsg_over_omega}
    \vspace{-4mm}
\end{figure}

\paragraph{Guidance Scale and Sparsity.}
\Cref{fig:maskdit_tread_combined,fig:routing_agsg_over_omega} vary the guidance scale $\omega$ alongside the sparsity controls $(\gamma_{\text{strong}}, \gamma_{\text{weak}})$. Across $\omega \in \{1.3, 1.5, 1.7, 1.9\}$ the optimal FID remains essentially unchanged, yet larger $\omega$ consistently tolerates higher total sparsity induced by $(\gamma_{\text{strong}}, \gamma_{\text{weak}})$. Consequently, jointly increasing $\omega$ and $(\gamma_{\text{strong}}, \gamma_{\text{weak}})$ improves efficiency while maintaining image quality. \Cref{fig:routing_agsg_over_omega} visualizes this with FID heatmaps whose color range is clipped to highlight the trend. The $(\gamma_{\text{strong}}, \gamma_{\text{weak}})$ valley shifts and steepens as $\omega$ increases. The optimum becomes more localized and flattens less while permitting higher sparsity. Intuitively, larger $\omega$ pairs well with higher inference-time sparsity because sparsity degrades the generated signal. This pushes samples farther from the target image manifold while stronger guidance scale $\omega$ counteracts this drift.

\paragraph{Routing vs.\ Masking.}
Routing withholds tokens temporarily and reinserts them unchanged, preserving instance-specific information and stabilizing guidance. Accordingly, the $(\gamma_{\text{strong}},\gamma_{\text{weak}})$ landscape is broader, supports higher total sparsity, and is less sensitive to hyperparameters. Masking entails irreversible token deletion but even in this regime SG remains effective. As expected, the response surface over $(\gamma_{\text{strong}}, \gamma_{\text{weak}})$ is narrower than that found in routing but a clear corridor achieves improved FID (see \Cref{fig:maskdit_tread_combined}). This demonstrates that even sparsities which intuitively do not align with the iterative refinement goal of diffusion, can still be used to effectively guide the model towards better quality using our proposed Sparse Guidance method.

\begin{figure}[b]
  \centering  
  \resizebox{\linewidth}{!}{
    \begin{minipage}{\linewidth}
      \centering
      \begin{subfigure}[b]{0.48\linewidth}
        \centering
        \includegraphics[width=\linewidth]{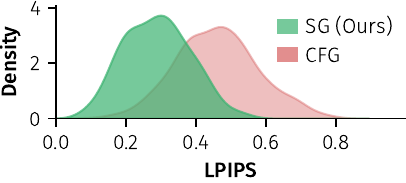}
        \label{fig:lpips_kde_overlay_sub}
      \end{subfigure}%
      \hspace{0.02\linewidth}
      \begin{subfigure}[b]{0.48\linewidth}
        \centering
        \includegraphics[width=\linewidth]{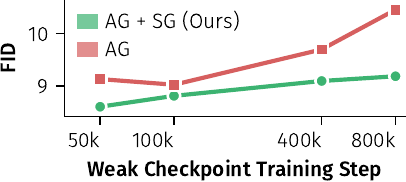}
        \label{fig:sg_ag_comparison_sub}
      \end{subfigure}
    \end{minipage}
  }
  \vspace{-10pt}
  \caption{\textbf{(Left) SG demonstrates smaller LPIPS between the output with guidance and the conditional prediction.} \textbf{(Right) SG allows for better usage of other, less flexible guidance methods, like AutoGuidance} by offering the capability to adjust network capacities without training for fine-grained capacity gaps.
  }
  \label{fig:sg_ag_combo}
\end{figure}

\paragraph{Compounding Gains with AutoGuidance.}
We further evaluate compatibility with external guidance by incorporating undertrained auxiliary models, following \citet{karras2024guiding}, within our Sparse Guidance (SG) framework. A central limitation of \emph{AutoGuidance} is the requirement for an additional training run with dense checkpointing: only a narrow window of auxiliary checkpoints yields high-quality results, and \citet{karras2024guiding} recommend dedicating $\tfrac{1}{16}$ of the total training iterations to the auxiliary model. This design is inherently inflexible, as the checkpoint cadence must be selected \emph{a priori}. In contrast, SG markedly relaxes these constraints. Instead of relying on a precise reference checkpoint, (near-) optimal auxiliary models can be recovered from a broad range of training steps by tuning the sparsity controls $\gamma_{\text{strong}}$ and $\gamma_{\text{weak}}$. As shown in \Cref{fig:sg_ag_combo}, we evaluate auxiliary checkpoints at 50k, 100k, 400k, and 800k steps—corresponding to 2.5\%, 5\%, 20\%, and 40\% of the total training iterations of $v_0$. For later checkpoints (800k and 400k), the best FID is achieved with $\gamma_{\text{strong}} = 0.0$. As we move to earlier checkpoints, the optimal $\gamma_{\text{strong}}$ for $v_0$ increases to preserve the relative gap between the $v_0$ and $v_1$ output distributions.
Overall, SG broadens the set of usable auxiliary checkpoints and compensates for their suboptimality through sparsity adaptation, delivering a favorable balance between FID and inference efficiency without committing to rigid checkpoint schedules.

\subsection{Sparse Guidance in Large-scale T2I Models} 
\begin{figure}[t]
  \centering
  \includegraphics[width=\linewidth]{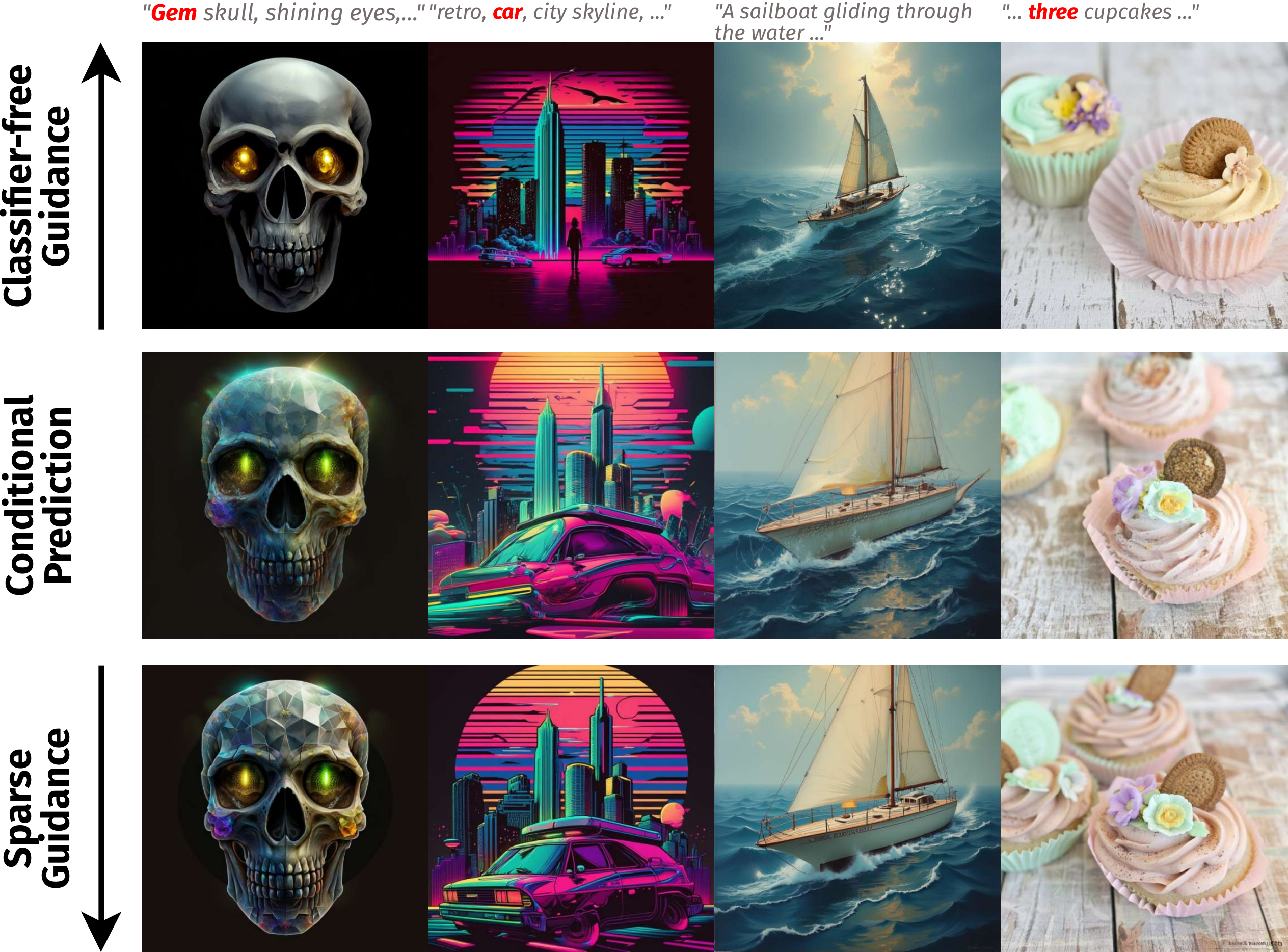}
  \caption{\textbf{Selected examples: Sparse Guidance keeps more of the structure of the conditional prediction} leading to higher variance in sample distribution while staying truthful to the prompt.}
  \label{fig:cfg_vs_sg_common_objects}
  \vspace{-4mm}
\end{figure}

To provide insights into a more complex task at scale, we train a 2.5B diffusion transformer with routing sparsity according to \citet{krause2025tread}. We evaluate our model using standard CFG and our proposed Sparse Guidance on common benchmarks like GenEval \cite{ghosh2023geneval} and HPSv3 \cite{ma2025hpsv3}. Instead of FID, we utilize HPSv3 as our metric of choice to determine sparsity rates $\gamma_{\text{strong}}$ and $\gamma_{\text{weak}}$. For this we use 250 synthetically generated prompts and the mean score over these.
Phenomena previously reported at small scale on ImageNet-256 also persist in our billion-parameter text-to-image setting: even without any guidance. \modelname's conditional branch exhibits clear, prompt- and layout-aware structure, consistent with the analysis of \citet{krause2025tread}. Furthermore, we confirm that Classifier-free Guidance (CFG) pulls the conditional predictor toward more stereotypical solutions. This aligns with the elevated \emph{Recall} we measure for SG in \Cref{tab:compact} and the qualitative trend in \Cref{fig:cfg_vs_sg_common_objects}.

\begin{table*}[t]
	\centering
	\adjustbox{max width=\linewidth}{
		\small
		\begin{tabular}{lrrrrrrrrrrrrrr}
			\toprule
			\textbf{Model}                                            & \textbf{Rank} $\downarrow$ & \textbf{Overall} $\uparrow$                      & \textbf{Characters}                   & \textbf{Arts}                         & \textbf{Design}                       & \textbf{Architecture}                 & \textbf{Animals}                      & \textbf{Natural Scenery}              & \textbf{Transportation}               & \textbf{Products}                     & \textbf{Others}                            & \textbf{Plants}                       & \textbf{Food}                         & \textbf{Science}                      \\
			\midrule
			\textcolor{lightgray}{Kolors \cite{kolors_2024}}  & \textcolor{lightgray}{1}                    & \textcolor{lightgray}{10.55} & \textcolor{lightgray}{11.79} & \textcolor{lightgray}{10.47} & \textcolor{lightgray}{9.87}  & \textcolor{lightgray}{10.82} & \textcolor{lightgray}{10.60} & \textcolor{lightgray}{9.89}  & \textcolor{lightgray}{10.68} & \textcolor{lightgray}{10.93} & \textcolor{lightgray}{10.50}      & \textcolor{lightgray}{10.63} & \textcolor{lightgray}{11.06} & \textcolor{lightgray}{9.51}  \\
			\textcolor{lightgray}{Flux-dev \cite{flux2024}} & \textcolor{lightgray}{2}                  & \textcolor{lightgray}{10.43} & \textcolor{lightgray}{11.70} & \textcolor{lightgray}{10.32} & \textcolor{lightgray}{9.39}  & \textcolor{lightgray}{10.93} & \textcolor{lightgray}{10.38} & \textcolor{lightgray}{10.01} & \textcolor{lightgray}{10.84} & \textcolor{lightgray}{11.24} & \textcolor{lightgray}{10.21}      & \textcolor{lightgray}{10.38} & \textcolor{lightgray}{11.24} & \textcolor{lightgray}{9.16}  \\
			\textcolor{lightgray}{Playgroundv2.5 \cite{playground_v2_5_2024}} & \textcolor{lightgray}{3}            & \textcolor{lightgray}{10.27} & \textcolor{lightgray}{11.07} & \textcolor{lightgray}{9.84}  & \textcolor{lightgray}{9.64}  & \textcolor{lightgray}{10.45} & \textcolor{lightgray}{10.38} & \textcolor{lightgray}{9.94}  & \textcolor{lightgray}{10.51} & \textcolor{lightgray}{10.62} & \textcolor{lightgray}{10.15}      & \textcolor{lightgray}{10.62} & \textcolor{lightgray}{10.84} & \textcolor{lightgray}{9.39}  \\
			\textcolor{lightgray}{Infinity \cite{Infinity}} & \textcolor{lightgray}{4}                  & \textcolor{lightgray}{10.26} & \textcolor{lightgray}{11.17} & \textcolor{lightgray}{9.95}  & \textcolor{lightgray}{9.43}  & \textcolor{lightgray}{10.36} & \textcolor{lightgray}{9.27}  & \textcolor{lightgray}{10.11} & \textcolor{lightgray}{10.36} & \textcolor{lightgray}{10.59} & \textcolor{lightgray}{10.08}      & \textcolor{lightgray}{10.30} & \textcolor{lightgray}{10.59} & \textcolor{lightgray}{9.62}  \\

            \rowcolor{gray!8}
            \tikzmarknode{sgW}{\coloruline{green!60!black}{\modelname{} + SG (Ours)}} &
            \tikzmarknode{rankSG}{\coloruline{green!60!black}{5}} &
            9.87 & 11.32 & 9.45 & 9.15 & 10.21 & 9.82 & 9.01 & 10.39 & 10.41 & 9.57 & 9.81 & 10.82 & \mk{sgE}{8.42} \\

			\textcolor{lightgray}{CogView4 \cite{cogview4_2025}} & \textcolor{lightgray}{6}                  & \textcolor{lightgray}{9.61}  & \textcolor{lightgray}{10.72} & \textcolor{lightgray}{9.86}  & \textcolor{lightgray}{9.33}  & \textcolor{lightgray}{9.88}  & \textcolor{lightgray}{9.16}  & \textcolor{lightgray}{9.45}  & \textcolor{lightgray}{9.69}  & \textcolor{lightgray}{9.86}  & \textcolor{lightgray}{9.45}       & \textcolor{lightgray}{9.49}  & \textcolor{lightgray}{10.16} & \textcolor{lightgray}{8.97}  \\
			\textcolor{lightgray}{PixArt-$\Sigma$ \cite{chen2024pixartsigma}} & \textcolor{lightgray}{7}           & \textcolor{lightgray}{9.37}  & \textcolor{lightgray}{10.08} & \textcolor{lightgray}{9.07}  & \textcolor{lightgray}{8.41}  & \textcolor{lightgray}{9.83}  & \textcolor{lightgray}{8.86}  & \textcolor{lightgray}{8.87}  & \textcolor{lightgray}{9.44}  & \textcolor{lightgray}{9.57}  & \textcolor{lightgray}{9.52}       & \textcolor{lightgray}{9.73}  & \textcolor{lightgray}{10.35} & \textcolor{lightgray}{8.58}  \\
			\textcolor{lightgray}{Gemini 2.0 Flash \cite{gemini_2_0_flash_2025}} & \textcolor{lightgray}{8}          & \textcolor{lightgray}{9.21}  & \textcolor{lightgray}{9.98}  & \textcolor{lightgray}{8.44}  & \textcolor{lightgray}{7.64}  & \textcolor{lightgray}{10.11} & \textcolor{lightgray}{9.42}  & \textcolor{lightgray}{9.01}  & \textcolor{lightgray}{9.74}  & \textcolor{lightgray}{9.64}  & \textcolor{lightgray}{9.55}       & \textcolor{lightgray}{10.16} & \textcolor{lightgray}{7.61}  & \textcolor{lightgray}{9.23}  \\

            \rowcolor{gray!8}
            \tikzmarknode{cfgW}{\coloruline{red}{\modelname{} + CFG}} &
            \tikzmarknode{rankCFG}{\coloruline{red}{9}} &
            9.21 & 10.54 & 9.33 & 9.15 & 9.34 & 9.41 & 8.44 & 9.36 & 9.51 & 8.57 & 9.34 & 10.42 & \tikzmarknode{cfgE}{8.60} \\

			\textcolor{lightgray}{Stable Diffusion XL \cite{podell2023sdxl}} & \textcolor{lightgray}{10}
              & \textcolor{lightgray}{8.20}  & \textcolor{lightgray}{8.67}  & \textcolor{lightgray}{7.63}  & \textcolor{lightgray}{7.53}  & \textcolor{lightgray}{8.57}  & \textcolor{lightgray}{8.18}  & \textcolor{lightgray}{7.76}  & \textcolor{lightgray}{8.65}  & \textcolor{lightgray}{8.85}  & \textcolor{lightgray}{8.32}       & \textcolor{lightgray}{8.43}  & \textcolor{lightgray}{8.78}  & \textcolor{lightgray}{7.29}  \\
			\textcolor{lightgray}{HunyuanDiT \cite{hunyuandit_2024}} & \textcolor{lightgray}{11}                & \textcolor{lightgray}{8.19}  & \textcolor{lightgray}{7.96}  & \textcolor{lightgray}{8.11}  & \textcolor{lightgray}{8.28}  & \textcolor{lightgray}{8.71}  & \textcolor{lightgray}{7.24}  & \textcolor{lightgray}{7.86}  & \textcolor{lightgray}{8.33}  & \textcolor{lightgray}{8.55}  & \textcolor{lightgray}{8.28}       & \textcolor{lightgray}{8.31}  & \textcolor{lightgray}{8.48}  & \textcolor{lightgray}{8.20}  \\

            \rowcolor{gray!8}
            \tikzmarknode{ungW}{\modelname{} (Unguided)} &
            \tikzmarknode{rankUng}{12} &
            7.76 & 8.49 & 8.04 & 8.33 & 7.97 & 6.63 & 7.77 & 7.40 & 7.38 & 7.02 & 8.02 & 8.06 & \tikzmarknode{ungE}{8.01} \\

			\textcolor{lightgray}{Stable Diffusion 3 Medium \cite{esser2024scalingrectifiedflowtransformers}} & \textcolor{lightgray}{13} & \textcolor{lightgray}{5.31}  & \textcolor{lightgray}{6.70}  & \textcolor{lightgray}{5.98}  & \textcolor{lightgray}{5.15}  & \textcolor{lightgray}{5.25}  & \textcolor{lightgray}{4.09}  & \textcolor{lightgray}{5.24}  & \textcolor{lightgray}{4.25}  & \textcolor{lightgray}{5.71}  & \textcolor{lightgray}{5.84}       & \textcolor{lightgray}{6.01}  & \textcolor{lightgray}{5.71}  & \textcolor{lightgray}{4.58}  \\
			\textcolor{lightgray}{Stable Diffusion 2 \cite{sd2_release_2022}} & \textcolor{lightgray}{14}                       & \textcolor{lightgray}{-0.24} & \textcolor{lightgray}{-0.34} & \textcolor{lightgray}{-0.56} & \textcolor{lightgray}{-1.35} & \textcolor{lightgray}{-0.24} & \textcolor{lightgray}{-0.54} & \textcolor{lightgray}{-0.32} & \textcolor{lightgray}{1.00}  & \textcolor{lightgray}{1.11}  & \textcolor{lightgray}{-0.01}      & \textcolor{lightgray}{-0.38} & \textcolor{lightgray}{-0.38} & \textcolor{lightgray}{-0.84} \\
			\bottomrule
		\end{tabular}

        \begin{tikzpicture}[remember picture,overlay]
          \coordinate (midUng) at ($(ungW.east)!0.8!(rankUng.west)$);
          \coordinate (midCfg) at ($(cfgW.east)!0.8!(rankCFG.west)$);
          \coordinate (midSg)  at ($(sgW.east)!0.8!(rankSG.west)$);
        
          \draw[-{Stealth[length=3mm]}, line width=0.9pt, red]
            (midUng)
              to[out=160, in=200, looseness=1.5]
                node[pos=0.6, xshift=8pt, text=red]{+2}
            (midCfg);
            
          \draw[-{Stealth[length=3mm]}, line width=0.9pt, green!60!black]
            (midUng)
              to[out=160, in=200, looseness=1.2]
                node[pos=0.7, xshift=8pt, text=green!60!black]{+5}
            (midSg);

        \end{tikzpicture}
	}

	\caption{\textbf{HPSv3 scores \cite{ma2025hpsv3} for our sparsely trained \modelname{}. SG improves over CFG in all categories} and enables our model to beat three additional models (\texttt{Gemini 2.0 Flash}, \texttt{PixArt-$\Sigma$} and \texttt{CogView4}). More precisely, our method improves sample quality by 27\% over the unguided model and 7\% over CFG while increasing throughput from 0.32 to 0.49 images/s on an H200 GPU.}
	\label{tab:model-scores}
    \vspace{-2mm}
\end{table*}

\paragraph{Visual Variance.}
Aside from oversaturation, CFG is known for variance-collapsing properties due to the fact that one extrapolates away from the unconditional signal in the direction of the conditional signal. While this is effective in overall image-prompt alignment, CFG can quickly produce similar looking images, especially with rare permutations on otherwise common objects (see \Cref{fig:cfg_vs_sg_common_objects}). Since Sparse Guidance utilizes token sparsity as a driving force for guidance, instead of the text conditioning, we find that it retains the high-variance, creative expressivity of the conditional prediction better. This is shown in \Cref{fig:teaser} and \Cref{fig:cfg_vs_sg_common_objects}.

\vspace{-2mm}
\paragraph{Performance Comparison.}
We evaluate \modelname on the GenEval benchmark \cite{ghosh2023geneval}, which assesses compositional text–image alignment across six categories: \emph{single object}, \emph{two objects}, \emph{counting}, \emph{colors}, \emph{relative position}, and \emph{color attribution}. GenEval uses off-the-shelf detectors and classifiers to verify prompt satisfaction. With a standard Classifier-free Guidance (CFG) setting, \modelname attains an overall score of 0.61. Incorporating our proposed SG method yields a score of 0.62, indicating a consistent improvement attributable to SG (see \Cref{tab:model_comparison}). SG improves performance in every category, evidencing a robust guidance signal for compositional grounding. Notably, on GenEval’s everyday-object prompts, where CFG already excels via variance-collapsing, prompt-faithful generation, SG still yields additional gains. We also show that our method can not only generate more correct images, as shown in GenEval, but also more visually appealing ones. In \Cref{tab:model-scores} we show HPSv3 scores taken from \citet{ma2025hpsv3}  and find that the addition of SG improves our model from matching \texttt{Gemini 2.0 Flash} to beating \texttt{CogView4} in overall score. In other words, SG allows our model to beat three additional models that it was previously not able to outperform. 

\begin{table}[t]
	\centering
	\adjustbox{max width=\linewidth}{
		\begin{tabular}{lccccccc}
			\toprule
			\textbf{Model} &
			\textbf{Overall} $\uparrow$ &
			\makecell{\textbf{Single}\\\textbf{object}} &
			\makecell{\textbf{Two}\\\textbf{object}} &
			\textbf{Counting} &
			\textbf{Colors} &
			\textbf{Position} &
			\makecell{\textbf{Color}\\\textbf{attribution}} \\
			\midrule
			\textcolor{lightgray}{Stable Diffusion v1.5 \cite{rombach2022high_latentdiffusion_ldm}}     & \textcolor{lightgray}{0.43} & \textcolor{lightgray}{0.97} & \textcolor{lightgray}{0.38} & \textcolor{lightgray}{0.35} & \textcolor{lightgray}{0.76} & \textcolor{lightgray}{0.04} & \textcolor{lightgray}{0.06} \\
			\textcolor{lightgray}{Stable Diffusion v2.1 \cite{sd2_release_2022}}     & \textcolor{lightgray}{0.50} & \textcolor{lightgray}{0.98} & \textcolor{lightgray}{0.51} & \textcolor{lightgray}{0.44} & \textcolor{lightgray}{0.85} & \textcolor{lightgray}{0.07} & \textcolor{lightgray}{0.17} \\
			\textcolor{lightgray}{Stable Diffusion XL \cite{podell2023sdxl}}       & \textcolor{lightgray}{0.55} & \textcolor{lightgray}{0.98} & \textcolor{lightgray}{0.74} & \textcolor{lightgray}{0.39} & \textcolor{lightgray}{0.85} & \textcolor{lightgray}{0.15} & \textcolor{lightgray}{0.23} \\
			\textcolor{lightgray}{PixArt-alpha \cite{chen2023pixartalphafasttrainingdiffusion}}              & \textcolor{lightgray}{0.48} & \textcolor{lightgray}{0.98} & \textcolor{lightgray}{0.50} & \textcolor{lightgray}{0.44} & \textcolor{lightgray}{0.80} & \textcolor{lightgray}{0.08} & \textcolor{lightgray}{0.07} \\
            \textcolor{lightgray}{Flux.1-dev \cite{flux2024}}                & \textcolor{lightgray}{0.66} & \textcolor{lightgray}{0.98} & \textcolor{lightgray}{0.79} & \textcolor{lightgray}{0.73} & \textcolor{lightgray}{0.77} & \textcolor{lightgray}{0.22} & \textcolor{lightgray}{0.45} \\
			\textcolor{lightgray}{DALL-E 3 \cite{betker2023improving}}                  & \textcolor{lightgray}{0.67} & \textcolor{lightgray}{0.96} & \textcolor{lightgray}{0.87} & \textcolor{lightgray}{0.47} & \textcolor{lightgray}{0.83} & \textcolor{lightgray}{0.43} & \textcolor{lightgray}{0.45} \\
			\textcolor{lightgray}{CogView4 \cite{cogview4_2025}}               & \textcolor{lightgray}{0.73} & \textcolor{lightgray}{0.99} & \textcolor{lightgray}{0.86} & \textcolor{lightgray}{0.66} & \textcolor{lightgray}{0.79} & \textcolor{lightgray}{0.48} & \textcolor{lightgray}{0.58} \\
			\textcolor{lightgray}{Stable Diffusion 3 Medium \cite{esser2024scalingrectifiedflowtransformers}}                & \textcolor{lightgray}{0.74} & \textcolor{lightgray}{0.99} & \textcolor{lightgray}{0.94} & \textcolor{lightgray}{0.72} & \textcolor{lightgray}{0.89} & \textcolor{lightgray}{0.33} & \textcolor{lightgray}{0.60} \\
			\textcolor{lightgray}{Janus-Pro-7B \cite{chen2025janus}}              & \textcolor{lightgray}{\textbf{0.80}} & \textcolor{lightgray}{0.99} & \textcolor{lightgray}{0.89} & \textcolor{lightgray}{0.59} & \textcolor{lightgray}{0.90} & \textcolor{lightgray}{\textbf{0.79}} & \textcolor{lightgray}{0.66} \\
			\midrule
			\modelname (Unguided)                            & 0.48 & 0.93 & 0.50 & 0.36 & 0.77 & 0.13 & 0.20 \\
			\modelname{} + CFG                               & 0.61 & 0.98 & 0.73 & 0.55 & 0.86 & 0.19 & 0.36 \\
			\modelname{} + SG                                & 0.62 & 0.99 & 0.73 & 0.55 & 0.87 & 0.20 & 0.39 \\

			\bottomrule
		\end{tabular}
	}
	\caption{\textbf{GenEval scores for our sparsely trained \modelname{}.} SG shows consistent improvements over CFG.}
	\label{tab:model_comparison}
    \vspace{-4mm}
\end{table}
\section{Conclusion}
Sparse training approaches for diffusion models have shown large improvements in recent years, but lacked adaption by the community as their performance and behavior during inference was unpredictable and weak. To overcome this, we propose Sparse Guidance (SG) which erases this issue and provides benefits like a higher variance in sampled outputs as well as fine control over the capacity gap driving guidance. With SG we achieve an  \textbf{FID  of 1.58} while reducing FLOPs by \textbf{25\%}, and can push to a \textbf{58\%} FLOPs reduction at performance on par with the dense SiT baseline. Then, we scale sparse training to 2.5B for a text-to-image task and find that SG holds up at scale, improving human preference score and increasing throughput. We hope our work encourages the community to explore token-sparse diffusion models for substantial savings in time, compute, and CO$_2$.

\section*{Acknowledgments}
This work has been supported by the Horizon Europe project ELLIOT (GA No. 101214398), the German Federal Ministry for Economic Affairs and Energy within the project “NXT GEN AI METHODS – Generative Methoden für Perzeption, Prädiktion und Planung”, the project “GeniusRobot” (01IS24083) funded by the Federal Ministry of Research, Technology and Space (BMFTR), and the BMWE ZIM-project (No. KK5785001LO4) “conIDitional LoRA”. The authors gratefully acknowledge the Gauss Center for Supercomputing for providing compute through the NIC on JUWELS/JUPITER at JSC and the HPC resources supplied by the NHR@FAU Erlangen.
We would like to thank Shih-Ying Yeh, Rami Seid, David Glukhov, and Swayam Bhanded for the insightful discussions. Further, we would like to thank Owen Vincent for continuous technical support.

{
    \small
    \bibliographystyle{ieeenat_fullname}
    \bibliography{main}
}

\clearpage
\setcounter{page}{1}

\newcommand{\rowpair}[1]{%
  \begin{tikzpicture}[baseline=(img.base)]
    \node[inner sep=0pt] (img)
      {\includegraphics[width=0.9\linewidth]{fig/suppl/rows/row_#1.pdf}};
    \node[anchor=east] at ([xshift=-0.6em,yshift=2.5em]img.west)
      {$\text{SG}_{\text{FLOPS}}$};
    \node[anchor=east] at ([xshift=-0.6em,yshift=-2.5em]img.west)
      {$\text{SG}_{\text{FID}}$};
  \end{tikzpicture}%
}

\newcommand{\twobytwogrid}[8]{%
  \begin{tikzpicture}[baseline=(img11.base)]
    \node[inner sep=0pt] (img11) at (0,0)
      {\includegraphics[width=0.23\linewidth]{#1}};
    \node[inner sep=0pt,anchor=west] (img12) at (img11.east)
      {\includegraphics[width=0.23\linewidth]{#2}};
    \node[inner sep=0pt,anchor=west] (img13) at (img12.east)
      {\includegraphics[width=0.23\linewidth]{#3}};
    \node[inner sep=0pt,anchor=west] (img14) at (img13.east)
      {\includegraphics[width=0.23\linewidth]{#4}};

    \node[inner sep=0pt,anchor=north] (img21) at (img11.south)
      {\includegraphics[width=0.23\linewidth]{#5}};
    \node[inner sep=0pt,anchor=west] (img22) at (img21.east)
      {\includegraphics[width=0.23\linewidth]{#6}};
    \node[inner sep=0pt,anchor=west] (img23) at (img22.east)
      {\includegraphics[width=0.23\linewidth]{#7}};
    \node[inner sep=0pt,anchor=west] (img24) at (img23.east)
      {\includegraphics[width=0.23\linewidth]{#8}};

    \node[anchor=east] at ([xshift=-0.8em]img11.west)
      {$\text{CFG}$};
    \node[anchor=east] at ([xshift=-0.8em]img21.west)
      {$\text{SG} {\text{ (Ours)}}$};
  \end{tikzpicture}%
}

\renewcommand\thefigure{A\arabic{figure}}
\renewcommand\thetable{A\arabic{table}}
\renewcommand\theequation{A\arabic{equation}}
\renewcommand{\thesection}{\Alph{section}} 
\setcounter{equation}{0}
\setcounter{table}{0}
\setcounter{figure}{0}
\setcounter{section}{0}

\maketitlesupplementary

\section{Implementation Details} \label{sec:implementation}

\vspace{-1mm}
\subsection{Training Details for T2I}

\paragraph{Architecture}
We implement our transformer models~\citep{vaswani2017attention,dosovitskiy2020image} largely following the Llama architecture~\citep{touvron2023llama}. In particular, we apply pre-normalization via RMSNorm~\citep{zhang2019root}, exclude bias parameters from all linear transformations, and employ rotary positional embeddings~\citep{su2024roformer} in an axial configuration following the approach of~\citet{crowson2024scalable}. The feedforward network (FFN) design mirrors that of Llama, utilizing the SwiGLU activation~\citep{shazeer2020glu} and an expansion ratio of $\frac{8}{3}$.

\paragraph{Model} We train a modern T2I diffusion transformer with 2.5B parameters. To apply TREAD \cite{krause2025tread}, we mask tokens and positional indices simultaneously and reintroduce them at layer 30. We use Internvl3-2B~\cite{zhu2025internvl3} as the text encoder. In addition, we incorporate insights from \citet{ma2024exploring}, specifically employing two TransformerLayers after the frozen VLM and using a general system prompt as a prefix to our captions: \texttt{"Describe the image by detailing the color, shape, size, texture, quantity, text, and spatial relationships of the objects."}. For more details on the model refer to \Cref{tab:hyperparam_t2i}.

\vspace{-3mm}
\paragraph{Data}
We use InternVL3-2B~\cite{zhu2025internvl3} to recaption a 100M-sample subset of COYO-700M~\cite{kakaobrain2022coyo-700m}, producing four captions per image. First, we generate a highly detailed description of the image and then progressively distill it into three additional levels: multi-sentence descriptions, single-sentence descriptions, and finally keyword-level summaries. For the last three, we use the language capacity of the VLM exclusively to cut down on cost. After a first training stage, we filter the COYO subset by aesthetics score (>5) and add synthetic data from JourneyDB \cite{pan2023journeydb} and Flux-6M \cite{fang2025flux}.

\begin{table}[h!]
\centering\small
\adjustbox{max width=\textwidth}{
\begin{tabular}{l c}
\toprule
\textbf{Hyperparameter} & \textbf{\modelname} \\
\midrule
\multicolumn{2}{l}{\textit{Optimizer}}\\
Batch size          & 3{,}072 \\
Optimizer           & AdamW \\
Learning rate       & $5\times10^{-5}$ \\
$(\beta_1,\beta_2)$ & (0.9, 0.95) \\
\midrule
\multicolumn{2}{l}{\textit{Architecture}}\\
Embedding dim       & 2{,}048 \\
Attention heads     & 16 \\
Transformer layers  & 34 \\
\midrule
\multicolumn{2}{l}{\textit{TREAD settings}}\\
Route       & $\mathbf{r}_{2\rightarrow30}$ \\
Selection ratio     & 0.5 \\
\bottomrule
\end{tabular}
}
\caption{Hyperparameter setup for our \modelname{} model and the TREAD routing schedule.}
\label{tab:hyperparam_t2i}
\end{table}

\subsection{Hyperparameters for ImageNet}
Unless stated otherwise we inherit the DiT~\citep{dit_peebles2022scalable} setting: AdamW \citep{loshchilov2017decoupled_adamw}, a fixed learning rate of $10^{-4}$, $(\beta_1,\beta_2)=(0.9,0.999)$, \texttt{bf16} precision, and latent-space training with the \texttt{stabilityai/sd-vae-ft-ema} VAE \citep{rombach2022high_latentdiffusion_ldm}.  When we finetune LR is dropped to $10^{-5}$. For routing and masking specific parameters refer to \Cref{tab:hyperparam}.

\begin{table}[h!]
\centering\small
\adjustbox{max width=\textwidth}{
\begin{tabular}{l c@{\hskip 20pt} c}
\toprule
\textbf{Hyperparameter} & \textbf{Routing} & \textbf{Masking} \\
\midrule
\multicolumn{3}{l}{\textit{Optimizer}}\\
Batch size          & 256 & 256 \\
Optimizer            & AdamW & AdamW \\
Learning rate        & $1\times10^{-4}$ & $1\times10^{-4}$ \\
$(\beta_1,\beta_2)$  & (0.9, 0.999) & (0.9, 0.999) \\
\midrule
\multicolumn{3}{l}{\textit{Finetune}}\\
Batch size          & 256 & 256 \\
Learning rate        & $1\times10^{-5}$ & $1\times10^{-5}$ \\
\midrule
\multicolumn{3}{l}{\textit{Architecture}}\\
Embedding dim        & 1{,}152 & 1{,}152 \\
Attention heads      & 16 & 16 \\
Transformer layers   & 28 & 28 \\
\midrule
\multicolumn{3}{l}{\textit{TREAD settings}}\\
Route        & $\mathbf{r}_{2\rightarrow24}$ & -- \\
Selection ratio      & 0.5 & -- \\
\midrule
\multicolumn{3}{l}{\textit{MaskDiT settings}}\\
$D^{\mathrm{dec}}$Embedding dim        & -- & 512 \\
$D^{\mathrm{dec}}$Attention heads      & -- & 16 \\
$D^{\mathrm{dec}}$Transformer layers   & -- & 8 \\
Selection ratio      & -- & 0.5 \\
\bottomrule
\end{tabular}
}
\caption{Hyperparameter setup for the XL/2 backbones with additional information for routing \cite{krause2025tread} and masking \cite{zheng2023fast_maskdit} methods. $D^{\mathrm{dec}}$ refers to the decoder head placed upon the normal DiT-XL/2. $\mathbf{r}_{2\rightarrow24}$ refers to the route from layer 2 to layer 24.}
\vspace{-2mm}
\label{tab:hyperparam}
\end{table}

\section{Experiment Details}
\subsection{Sparse Guidance in ImageNet}
\paragraph{$\text{SG}_{\text{FLOPS}}$} from \Cref{tab:extensive_reduced} is obtained using the same checkpoint for the high capacity and low capacity model. Both are conditional and the distribution discrepancy is created solely via different routing rates. We find $\gamma_{\text{strong}}=0.5, \gamma_{\text{weak}}=0.9$ to achieve good FID while substantially decreasing FLOPS.

\paragraph{$\text{SG}_{\text{FID}}$}  (see \Cref{tab:extensive_reduced}, \Cref{tab:compact}) is obtained through the usage of an early checkpoint of the same model training run. More specifically, we utilize a checkpoint with 50k training iterations. Furthermore, we apply cosine decay from 0.6 to 0.0 on the auxiliary model and the inverse on the main model. This aligns with the findings from \Cref{fig:sg_ag_combo} where $\gamma_{\text{strong}}, \gamma_{\text{weak}}$ can be used to make up for undertrained auxiliary models. We achieve similar FID with other checkpoints and adjusted routing rates. 

\subsection{Sparse Guidance in Large Scale T2I Models}
\begin{figure}
    \centering
    \includegraphics[width=1.0\linewidth]{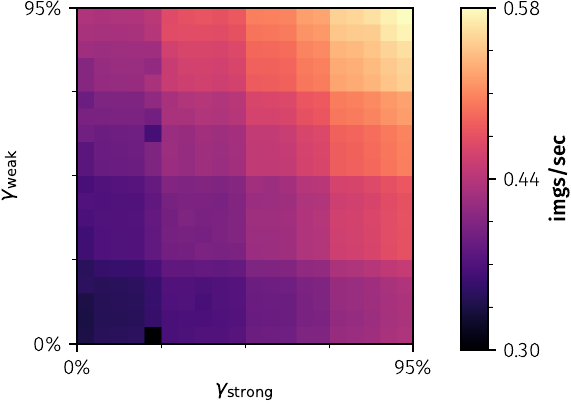}
    \caption{Inference speed for the guided setting. Lower left corner with zero $\gamma_{\text{strong}}$, $\gamma_{\text{weak}}$ resembles naive guided inference. Introducing sparsity (Sparse Guidance) allows for drastically improved throughput showcased by brighter colors towards the top right corner.}
    \label{fig:img_p_second_t2i_model}
\end{figure}

In \Cref{tab:model-scores} we show that applying our proposed Sparse Guidance to scaled T2I models yields better performance than CFG. Additionally, Sparse Guidance enables faster inference as seen in \Cref{fig:img_p_second_t2i_model} where a grid over the $\gamma_{\text{strong}}$,$\gamma_{\text{weak}}$ with a $0.05$ stepsize is shown.

\paragraph{GenEval \cite{ghosh2023geneval}} For GenEval (see \Cref{tab:model_comparison}), we stack our proposed Sparse Guidance method on top of Classifier-free Guidance and utilize $\omega=2.5, \gamma_{\text{strong}}=0.2$ and $\gamma_{\text{weak}}=0.7$.

\paragraph{HPSv3 \cite{ma2025hpsv3}} For the HPSv3 score (see \Cref{tab:model-scores}), we follow the proposed benchmark in \citet{ma2025hpsv3} with identical prompts. We utilize Sparse Guidance with $\omega=1.8, \gamma_{\text{strong}}=0.1$ and $\gamma_{\text{weak}}=0.8$.

\section{Auxiliary MAE loss under Flow Matching}
To facilitate a fair comparison between our SiT \cite{ma2024sit} baseline and MaskDiT \cite{zheng2023fast_maskdit}, we derive the MaskedAutoEncoder (MAE) loss for the flow-matching objective (see \Cref{tab:masking_vs_routing}, \Cref{fig:maskdit_tread_combined}).
MaskDiT~\cite{zheng2023fast_maskdit} combines a score-matching loss on visible tokens with a masked reconstruction (MAE) objective on masked tokens in diffusion models.
We generalize this formulation to the \emph{flow-matching} objective. Let $\mathcal{I}$ denote the token index set and $\mathbf{M}\in\{0,1\}^{\mathcal{I}}$ a random binary mask ($1$ for masked, $0$ for visible).
We define the visible mask as $\bar{\mathbf{M}}=\mathbf{1}-\mathbf{M}$. Following~\cite{zheng2023fast_maskdit}, the masked reconstruction loss is:
\begin{equation}
{
\footnotesize
\begin{aligned}
\mathcal{L}_{\text{MAE}} =\mathbb{E}_{x\sim p_{\text{data}}}\mathbb{E}_{t\sim[0,1]} \mathbb{E}_{\mathbf{M}} \big\|\big( D_{\theta} (x_t \odot \bar{\mathbf{M}} ,t) - x\big) \odot \mathbf{M}\big\|^2,
\end{aligned}
\label{eq:mae-diffusion}
}
\end{equation}
where $D_{\theta}$ predicts the denoised image at time~$t$ and $\odot$ denotes the Hadamard product. Unlike diffusion models, which predict the score~$\nabla_{x_t}\log p_t(x_t)$, flow matching directly parameterizes the instantaneous displacement of particles along this trajectory.
Given the path definition in Eq.~\ref{eq:path}, the latent states satisfy
\begin{equation}
\label{eq:residual-id}
\begin{aligned}
x - x_t \;=\; (1-t)(x - z) \;=\; (1-t)\,v^\star(x_{t},t),
\end{aligned}
\end{equation}
where $v^\star(x_{t},t)$ is the oracle velocity field driving the transformation from~$z$ to~$x$.
This relation reveals that reconstructing a future state~$x_{t}$ from a clean sample~$x$ is equivalent to estimating the target velocity~$v^\star(x_{t},t)$ up to the scalar factor~$(1 - t)$. Hence, in the flow-matching formulation, masked reconstruction can be interpreted as learning to predict the intermediate flow direction that transports partially visible tokens toward their clean targets. Replacing $v^\star$ by its learned approximation $v_\theta$, we have
\[
D_\theta(x_{t},t)-x_{t}\approx (1-t)\,v_\theta(x_{t},t).
\]
Consequently, the masked reconstruction term restricted to masked tokens can be reformulated as:
\begin{equation}
{
\footnotesize
\begin{aligned}
\mathcal{L}_{\text{MAE}} &= \mathbb{E}_{x}\mathbb{E}_{t\sim[0,1]} \mathbb{E}_{\mathbf{M}} 
\big\|(1-t)\, v_\theta(x_{t} \odot \bar{\mathbf{M}},{t}) \odot \mathbf{M}\big\|^2\\
&=\mathbb{E}_{x}\mathbb{E}_{t\sim[0,1]} \mathbb{E}_{\mathbf{M}} 
(1-t)^2 \big\|v_\theta(x_{t} \odot \bar{\mathbf{M}},{t}) \odot \mathbf{M}\big\|^2.
\end{aligned}
\label{eq:mae-diffusion-fm}
}
\end{equation}

The overall training objective combines the standard flow-matching loss with the auxiliary masked reconstruction term. According to \cite{krause2025tread}, routing models do not require additional auxiliary losses, so we use the standard flow matching objective. The final loss is defined as
\begin{equation}
{
\footnotesize
\label{eq:fm-mask}
\begin{aligned}
\mathcal{L}_{\mathrm{FM\text{-}mask}} &=
\mathbb{E}_{x,z,t}\Big[
\big\|\,\bar{\mathbf{M}}\odot\!\big(v_\theta(x_t,t)-v^\star(x_t,t)\big)\,\big\|_2^2 \\
&\quad
+ \lambda\,\mathbb{E}_{x, t,\mathbf{M}} (1-t)^2 \big\|v_\theta(x_{t} \odot \bar{\mathbf{M}},{t}) \odot \mathbf{M}\big\|_2^2\Big],
\end{aligned}
}
\end{equation}
where $\lambda$ balances the contribution of the masked reconstruction objective.
In practice, we set $\lambda$ empirically to ensure comparable magnitudes of the gradient between the two terms.

\section{Guidance Interaction Exploration}
Using the notation from Sec.~3.2, let
\[
D_s(c) := D_\theta(x_t,t,c;\gamma_{\text{strong}}),
\qquad
D_w(c) := D_\theta(x_t,t,c;\gamma_{\text{weak}}),
\]
with analogous definitions for \(D_s(\emptyset)\) and
\(D_w(\emptyset)\). We compare four interaction strategies
for combining CFG and SG.

\paragraph{A) Direct.}
We directly add the CFG direction and the SG direction:
\begin{equation}
\begin{aligned}
\tilde D_\theta^{\text{direct}}
&=
D_w(\emptyset)
+
w_{\text{cfg}}
\bigl(
D_s(c)-D_w(\emptyset)
\bigr) \\
&\quad +
w_{\text{sg}}
\bigl(
D_s(c)-D_w(c)
\bigr).
\end{aligned}
\end{equation}

\paragraph{B) Inner.}
Following the Inner-Guidance form, we use the jointly
conditioned prediction and subtract the partially dropped
branches:
\begin{equation}
\begin{aligned}
\tilde D_\theta^{\text{inner}}
&=
(1+w_{\text{cfg}}+w_{\text{sg}})\,D_s(c) \\
&\quad
-
w_{\text{cfg}}\,D_s(\emptyset)
-
w_{\text{sg}}\,D_w(c).
\end{aligned}
\end{equation}

\paragraph{C) Compositional.}
Following compositional guidance, we combine the
condition-specific predictions additively around a shared
unconditional branch:
\begin{equation}
\begin{aligned}
\tilde D_\theta^{\text{comp}}
&=
D_w(\emptyset)
+
w_{\text{cfg}}
\bigl(
D_s(c)-D_w(\emptyset)
\bigr) \\
&\quad +
w_{\text{sg}}
\bigl(
D_w(c)-D_w(\emptyset)
\bigr).
\end{aligned}
\end{equation}

\paragraph{D) IP2P-style.}
Following InstructPix2Pix, we apply the two signals
sequentially, first SG and then CFG:
\begin{equation}
\begin{aligned}
\tilde D_\theta^{\text{IP2P}}
&=
D_w(\emptyset)
+
w_{\text{sg}}
\bigl(
D_w(c)-D_w(\emptyset)
\bigr) \\
&\quad +
w_{\text{cfg}}
\bigl(
D_s(c)-D_w(c)
\bigr).
\end{aligned}
\end{equation}

Among these interaction strategies, the direct formulation
performs best, achieving the lowest FID while also requiring
the fewest function evaluations per step.

\begin{table}[h]
	\centering
    \vspace{-2mm}
	\begin{adjustbox}{max width=0.62\linewidth}
		\begin{tabular}{l@{\hskip .2em}c@{\hskip .2em}c}
			\toprule
			\textbf{Interaction Method} & \textbf{FID}$\downarrow$ & \textbf{NFE/step}$\downarrow\!$ \\
			\midrule
			A) direct                  & \textbf{2.14}            & \textbf{2}           \\
			B) inner                   & 10.70                    & 3 \\
			C) compositional           & 11.14                    & 3 \\
			D) IP2P-style              & 2.62                     & 3 \\
			\bottomrule
		\end{tabular}
	\end{adjustbox}
    \vspace{-3mm}
    \caption{Comparison of guidance interaction strategies on ImageNet.}
    \label{tab:model_comparison_combination}
    \vspace{-4mm}
\end{table}

\section{Qualitative Samples}
\begin{figure*}[t]
  \centering
  \twobytwogrid
    {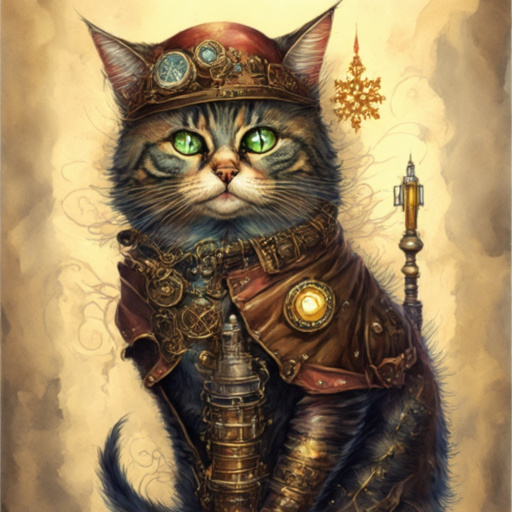}
    {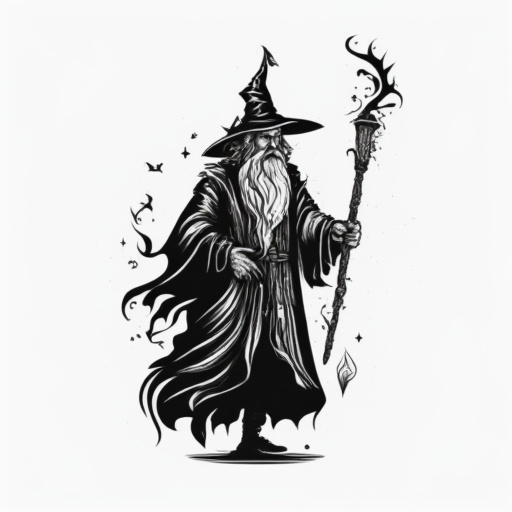}
    {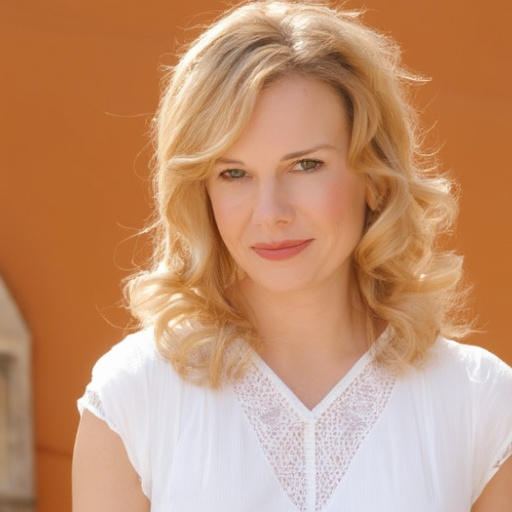}
    {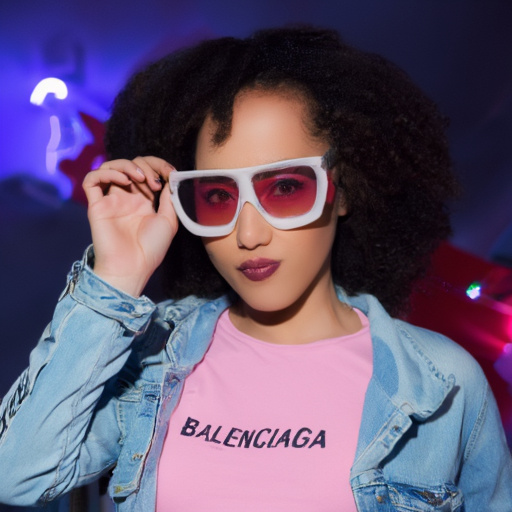}
    {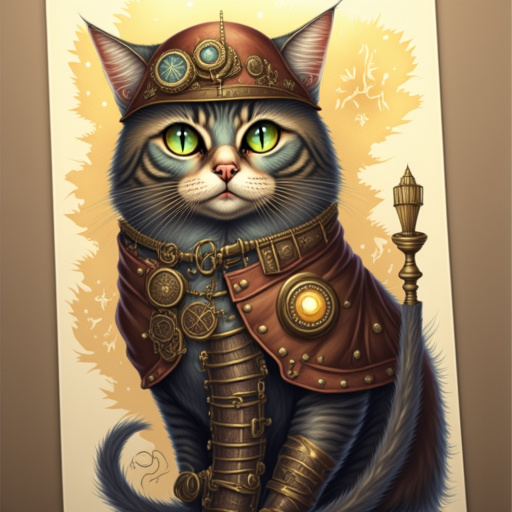}
    {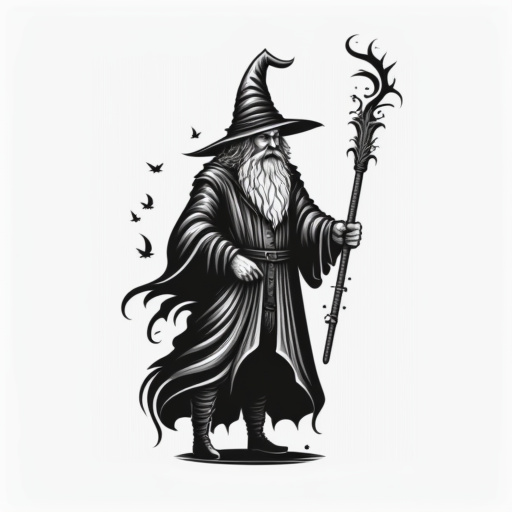}
    {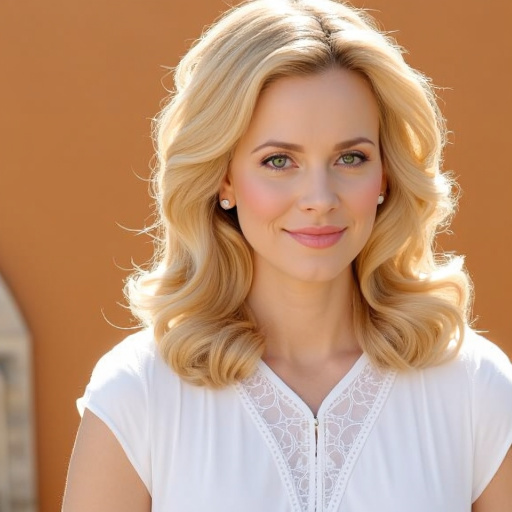}
    {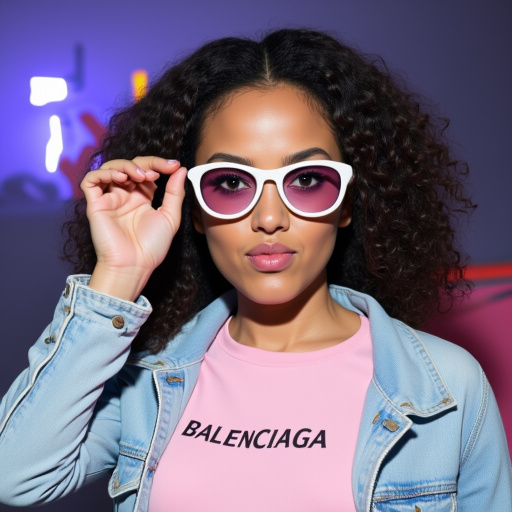}
    \caption{Qualitative examples comparing CFG to our proposed SG. Images with CFG tend to have more artifacts or seem blurry. SG provides crisp images with lower cost.}
  \label{fig:twobytwo_grid}
\end{figure*}

\begin{figure*}[t]
  \centering
  \twobytwogrid
    {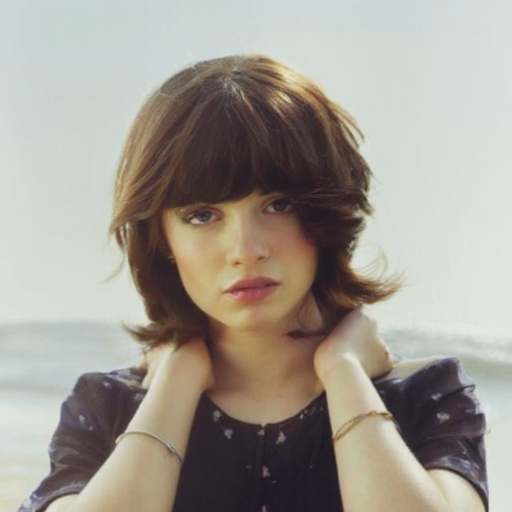}
    {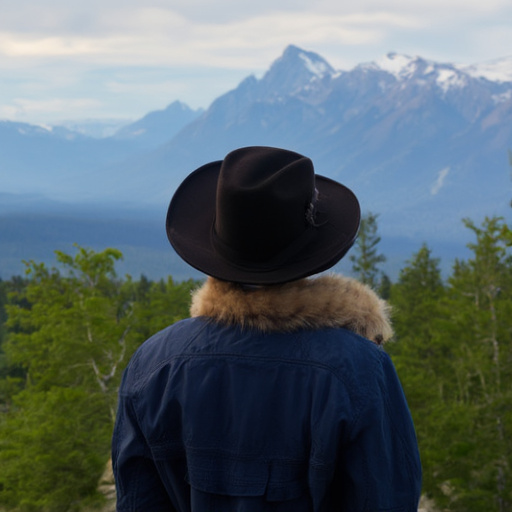}
    {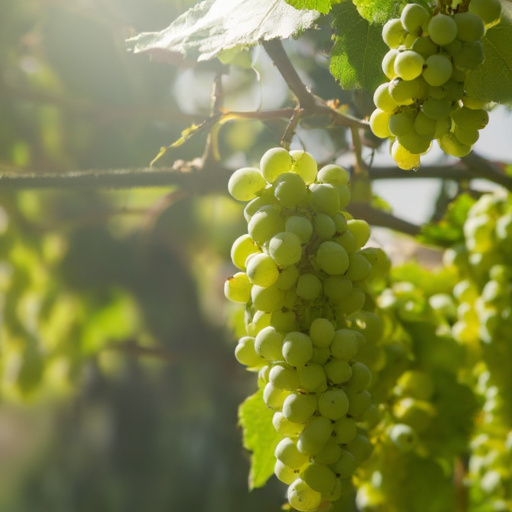}
    {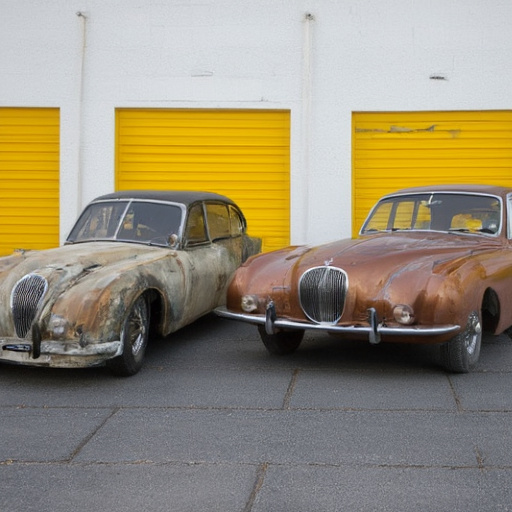}
    {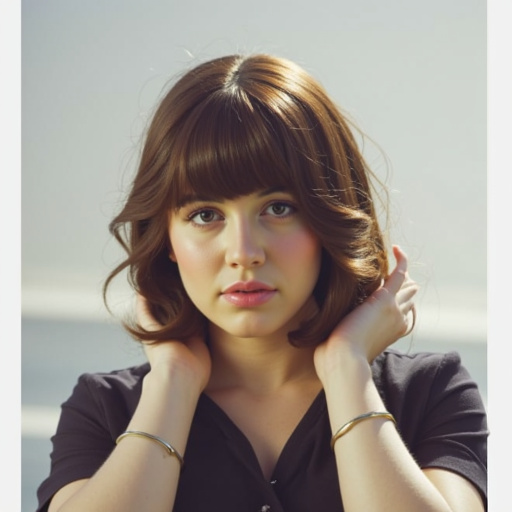}
    {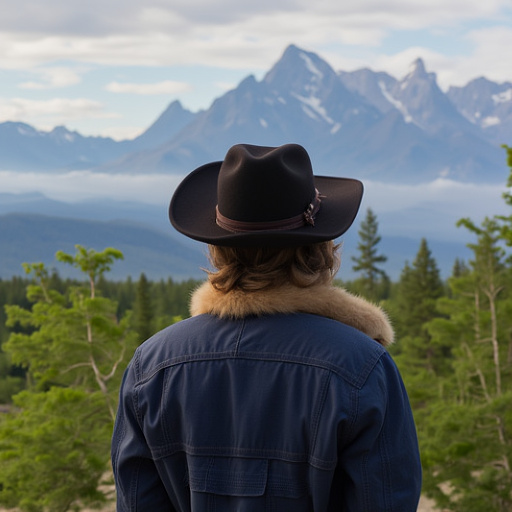}
    {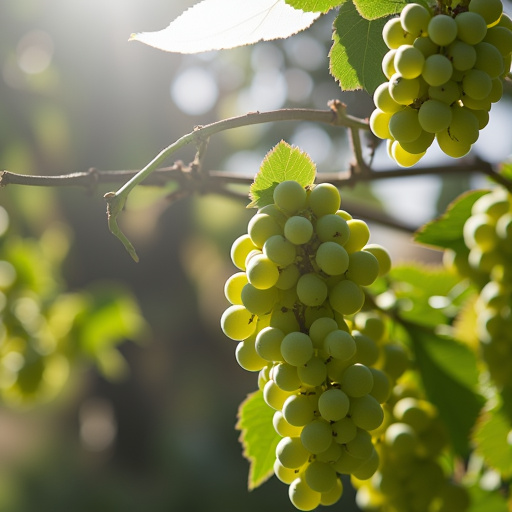}
    {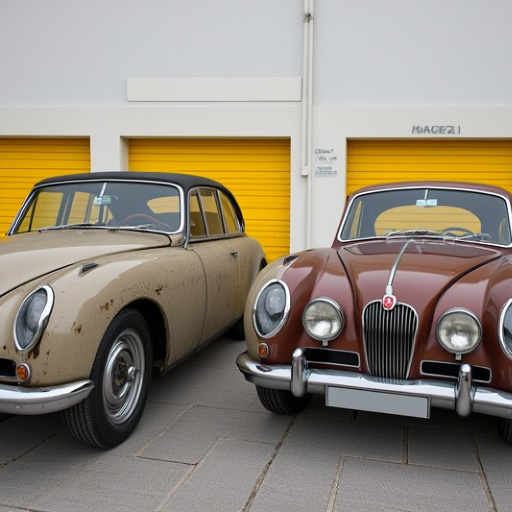}
    \caption{Qualitative examples comparing CFG to our proposed SG. Images with CFG tend to have more artifacts or seem blurry. SG provides crisp images with lower cost.}
  \label{fig:twobytwo_grid2}
\end{figure*}

\begin{figure*}[p]
    \centering
    \setlength{\tabcolsep}{0pt}
    \renewcommand{\arraystretch}{0}

    \begin{tabular}{cccccc}
        \includegraphics[width=.16\textwidth]{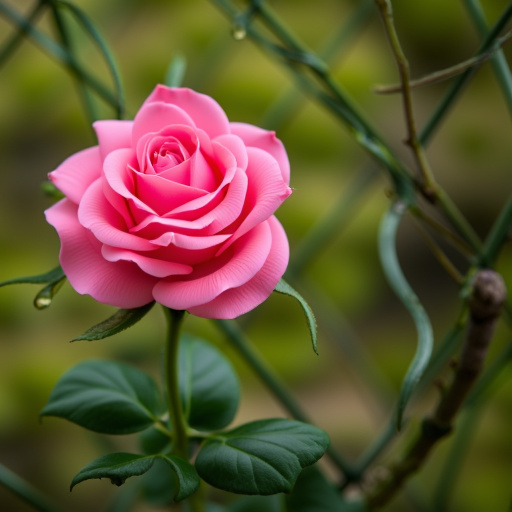} &
        \includegraphics[width=.16\textwidth]{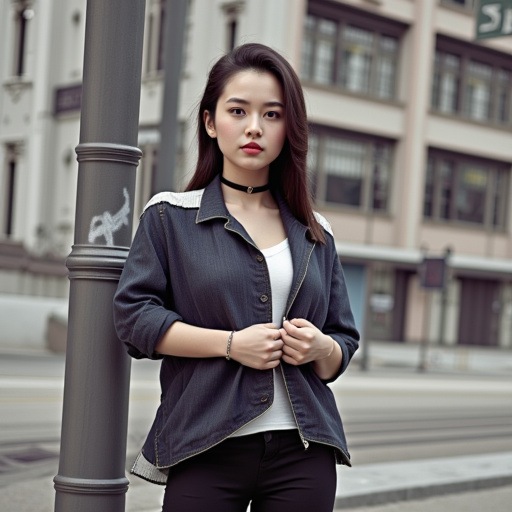} &
        \includegraphics[width=.16\textwidth]{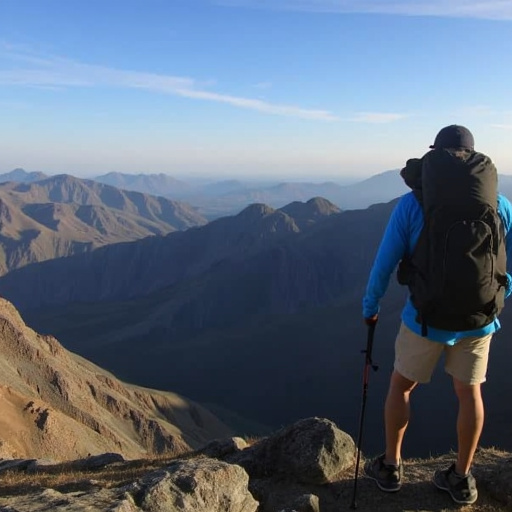} &
        \includegraphics[width=.16\textwidth]{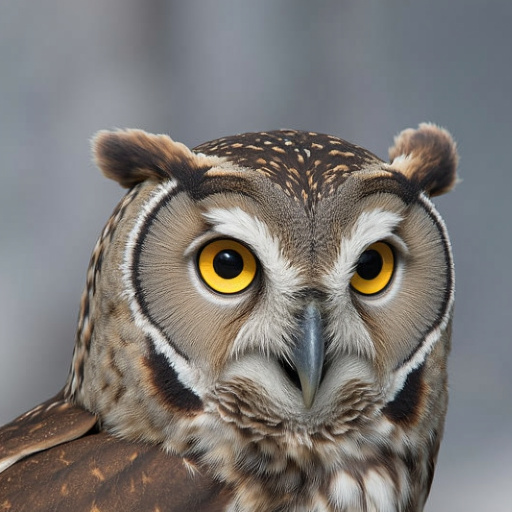} &
        \includegraphics[width=.16\textwidth]{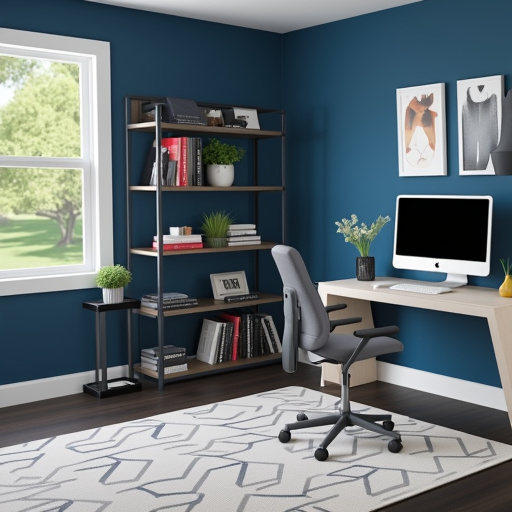} &
        \includegraphics[width=.16\textwidth]{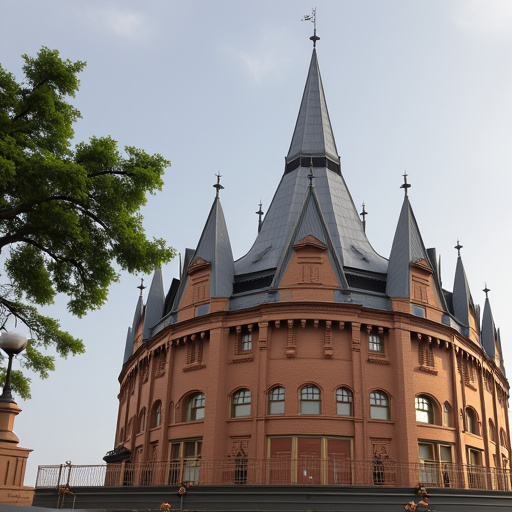} \\
        \includegraphics[width=.16\textwidth]{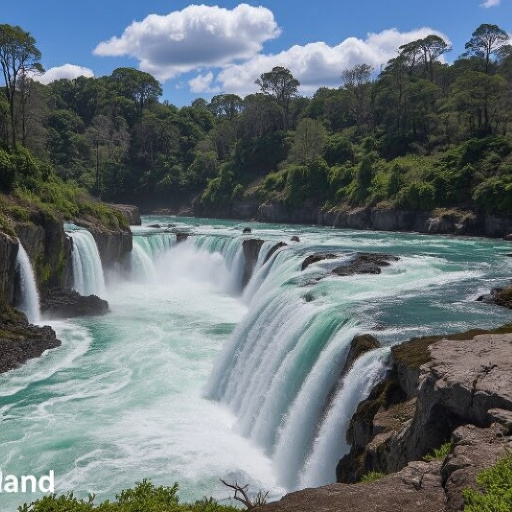} &
        \includegraphics[width=.16\textwidth]{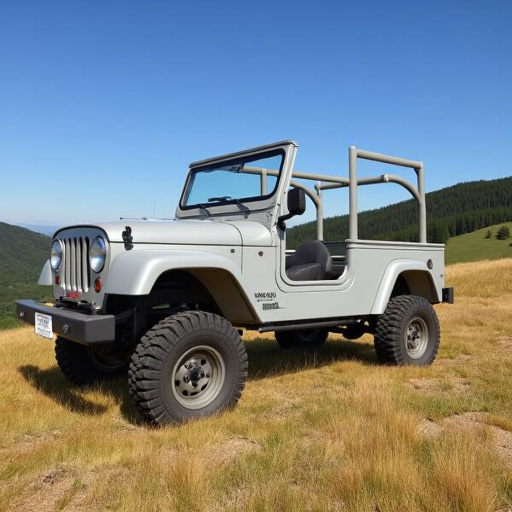} &
        \includegraphics[width=.16\textwidth]{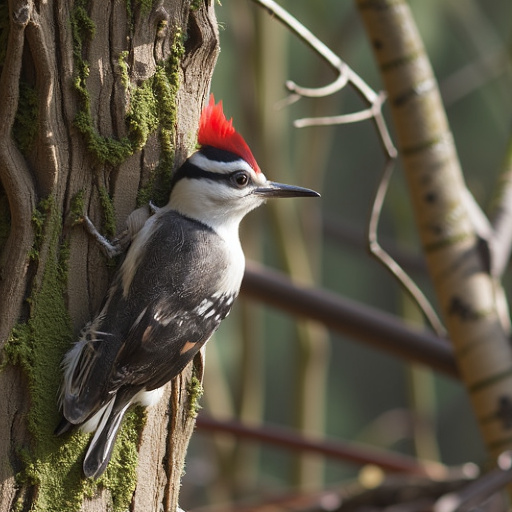} &
        \includegraphics[width=.16\textwidth]{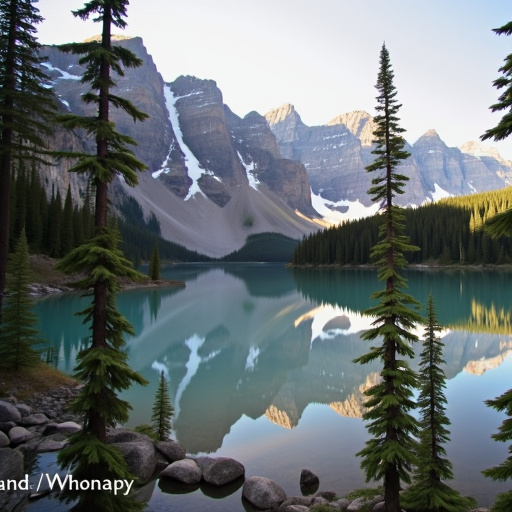} &
        \includegraphics[width=.16\textwidth]{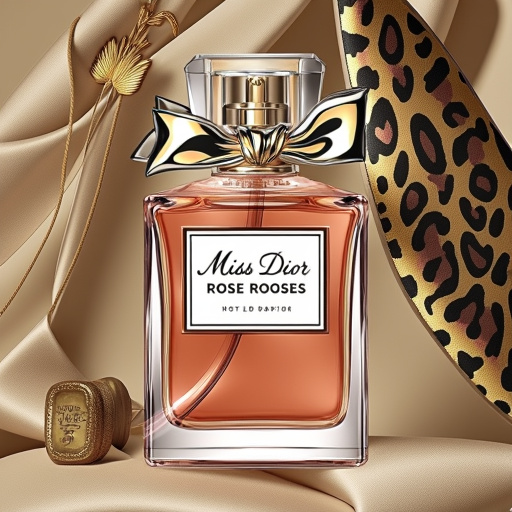} &
        \includegraphics[width=.16\textwidth]{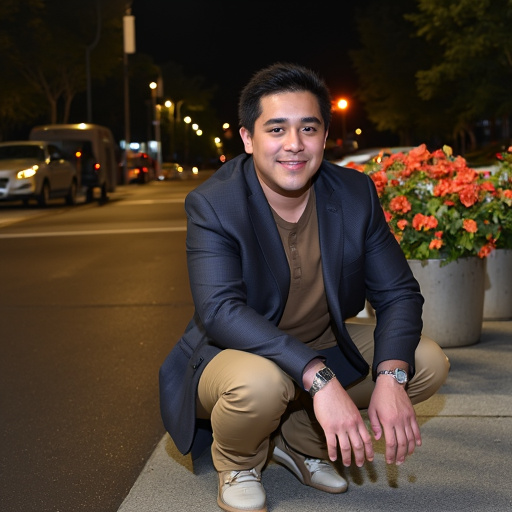} \\
        \includegraphics[width=.16\textwidth]{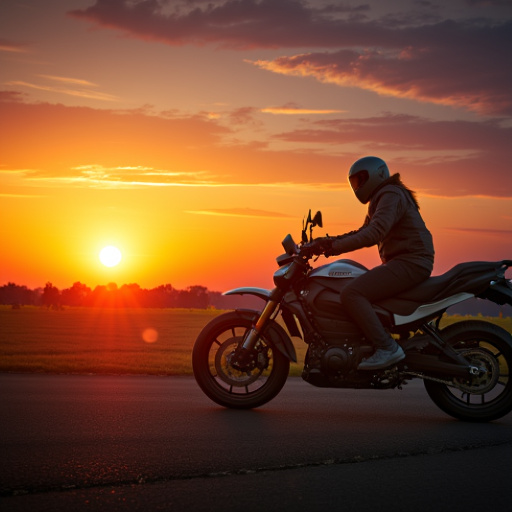} &
        \includegraphics[width=.16\textwidth]{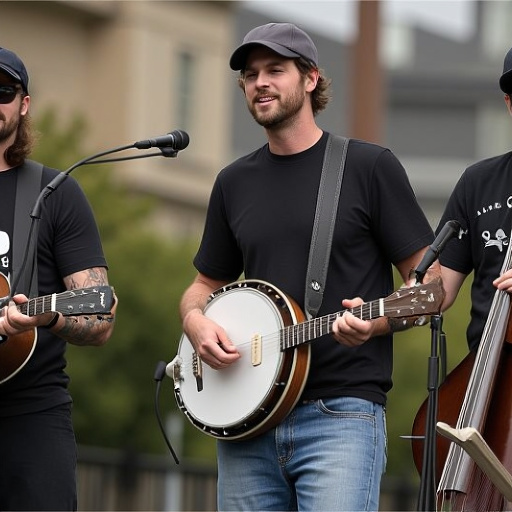} &
        \includegraphics[width=.16\textwidth]{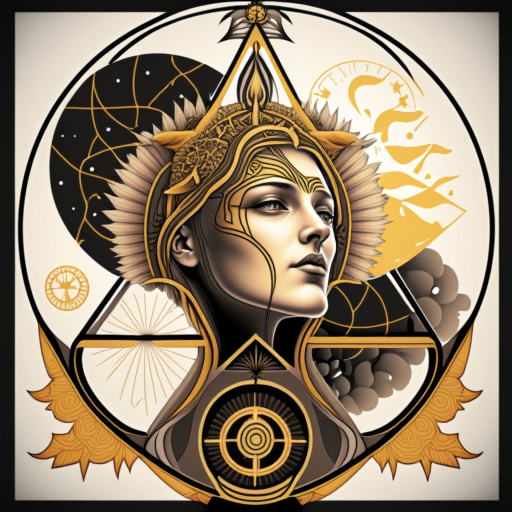} &
        \includegraphics[width=.16\textwidth]{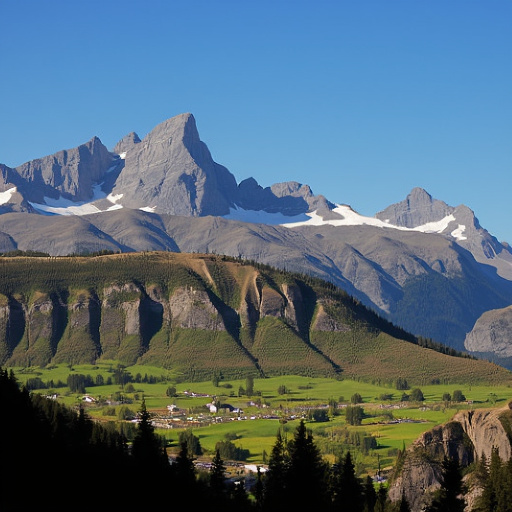} &
        \includegraphics[width=.16\textwidth]{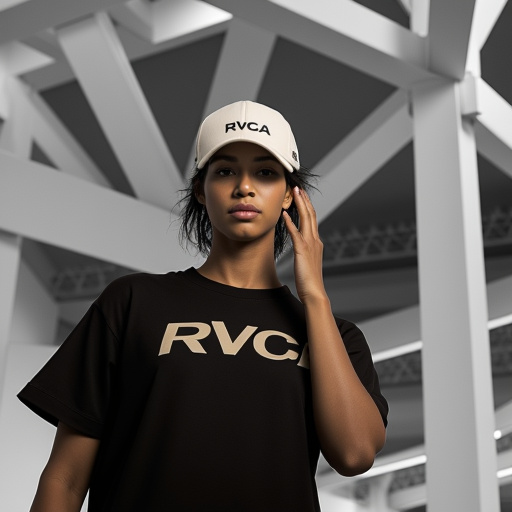} &
        \includegraphics[width=.16\textwidth]{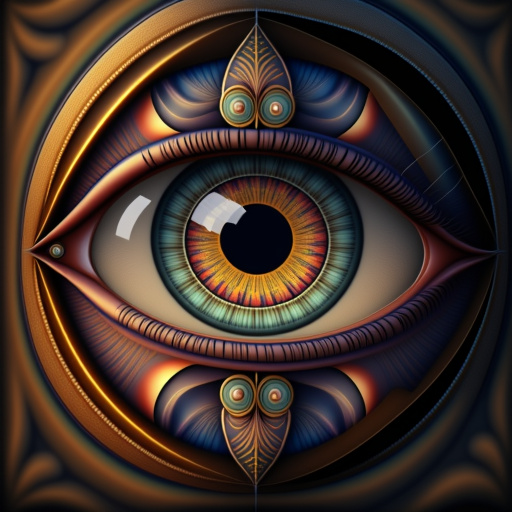} \\
        \includegraphics[width=.16\textwidth]{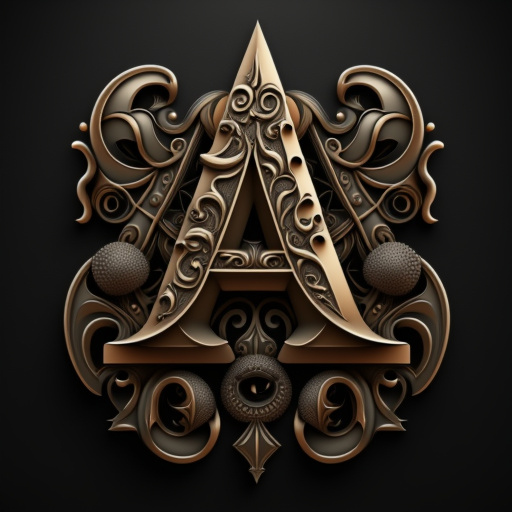} &
        \includegraphics[width=.16\textwidth]{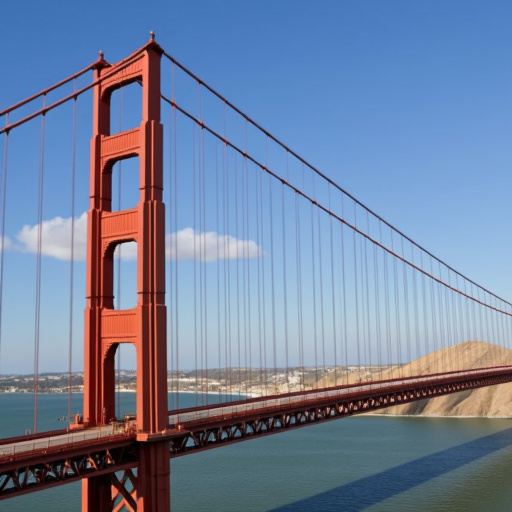} &
        \includegraphics[width=.16\textwidth]{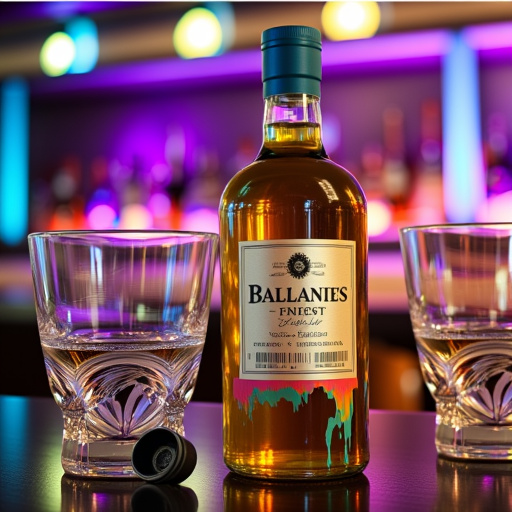} &
        \includegraphics[width=.16\textwidth]{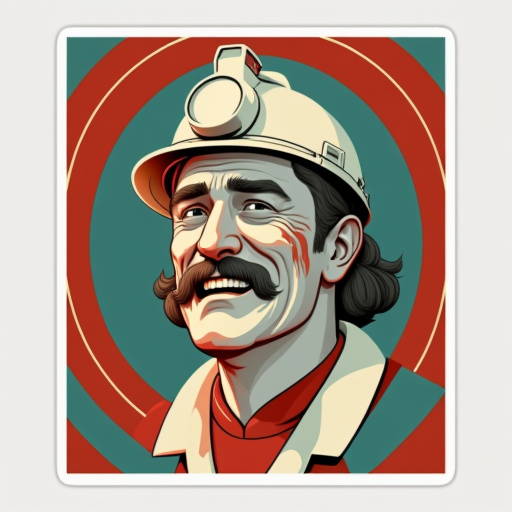} &
        \includegraphics[width=.16\textwidth]{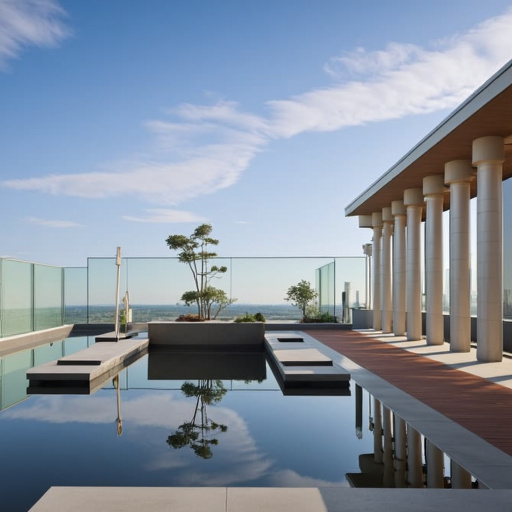} &
        \includegraphics[width=.16\textwidth]{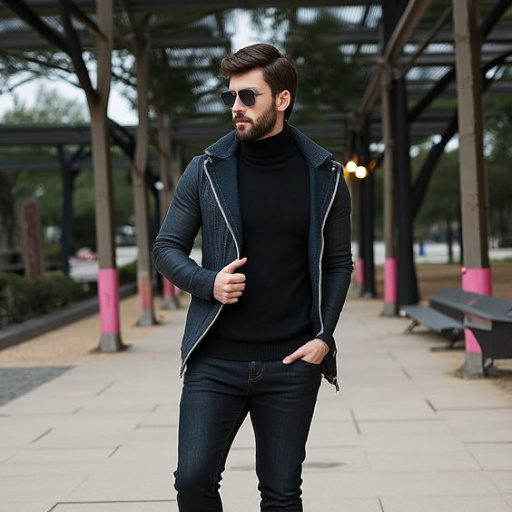} \\
        \includegraphics[width=.16\textwidth]{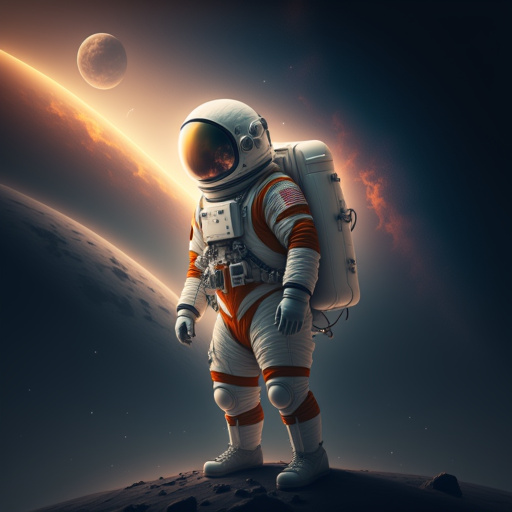} &
        \includegraphics[width=.16\textwidth]{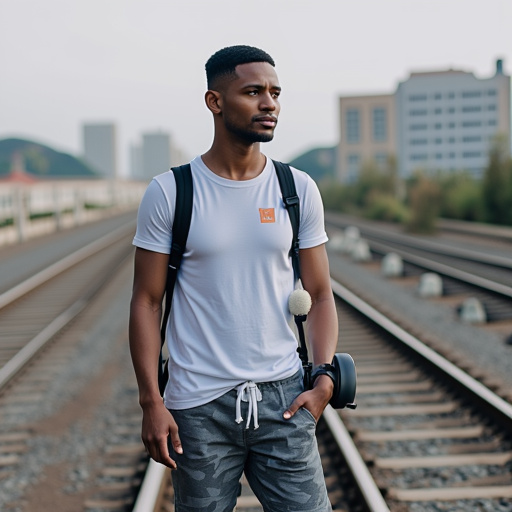} &
        \includegraphics[width=.16\textwidth]{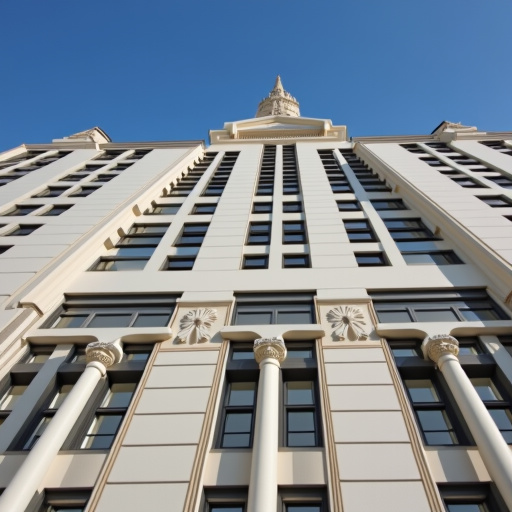} &
        \includegraphics[width=.16\textwidth]{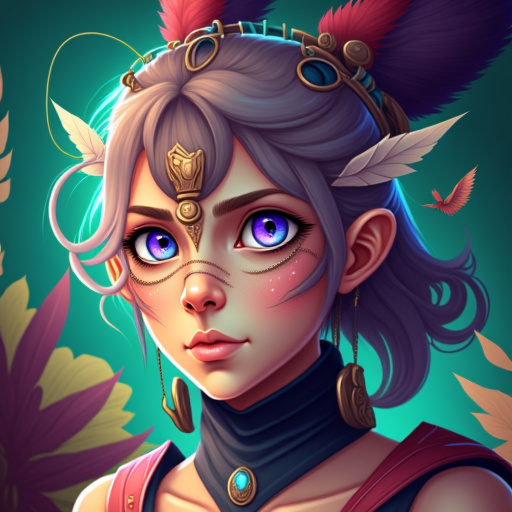} &
        \includegraphics[width=.16\textwidth]{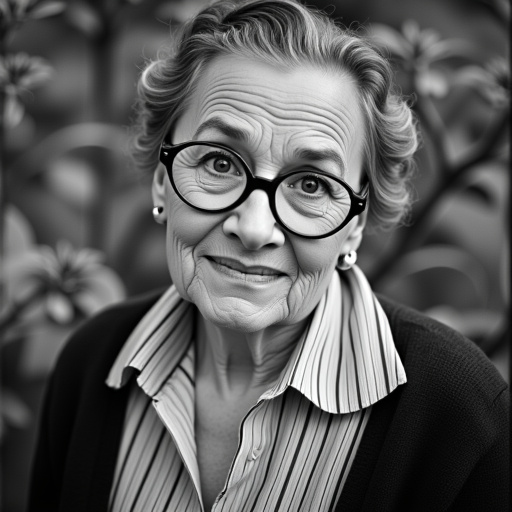} &
        \includegraphics[width=.16\textwidth]{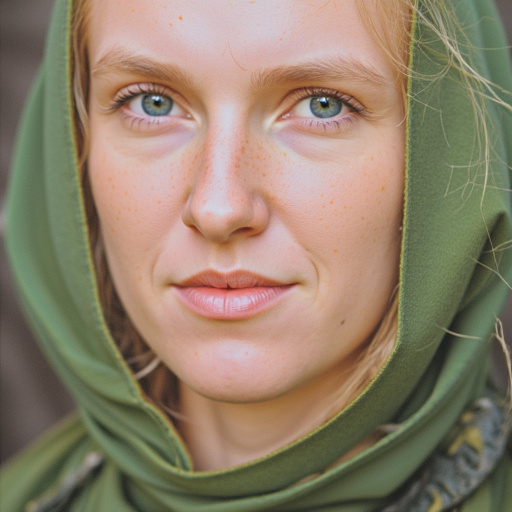} \\
    \end{tabular}

    \caption{Additional samples generated using Sparse Guidance. Prompts are taken from the HPSv3 benchmark subset.}
    \label{fig:sg_additional_30_5x6}
\end{figure*}

We provide additional qualitative text-to-image results in \Cref{fig:twobytwo_grid} and \Cref{fig:twobytwo_grid2}, where we directly compare Classifier-Free Guidance (CFG) with Sparse Guidance (SG) in our \modelname. Complementing these comparisons, \Cref{fig:sg_additional_30_5x6} presents a broader selection of SG-generated outputs. All text-to-image samples are produced using prompts sourced from the HPSv3~\cite{ma2025hpsv3} benchmark subset.

\begin{figure*}[h]
  \centering
  \setlength{\tabcolsep}{1pt}
  \begin{tabular}{cccc}
    \textbf{Unguided} & \textbf{SG} & \textbf{AG} & \textbf{CFG} \\[2pt]

    \includegraphics[width=.24\textwidth]{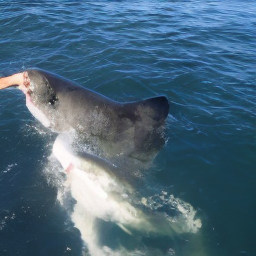} &
    \includegraphics[width=.24\textwidth]{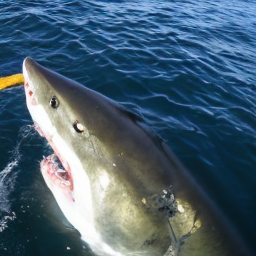}   &
    \includegraphics[width=.24\textwidth]{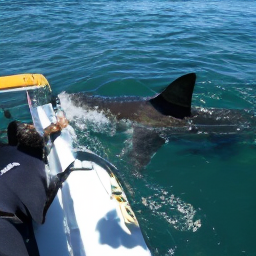}   &
    \includegraphics[width=.24\textwidth]{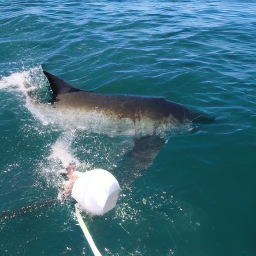}  \\[2pt]

    \includegraphics[width=.24\textwidth]{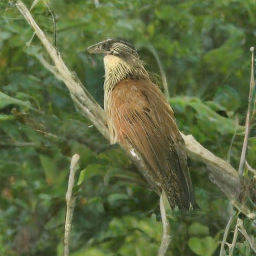} &
    \includegraphics[width=.24\textwidth]{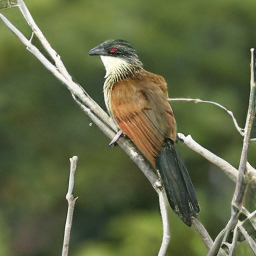}   &
    \includegraphics[width=.24\textwidth]{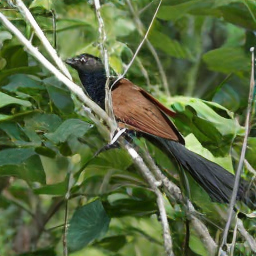}   &
    \includegraphics[width=.24\textwidth]{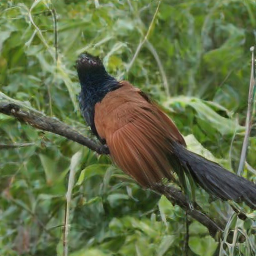}  \\[2pt]

    \includegraphics[width=.24\textwidth]{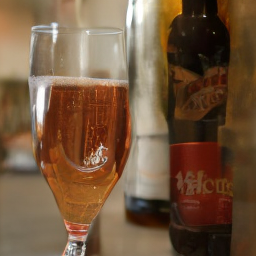} &
    \includegraphics[width=.24\textwidth]{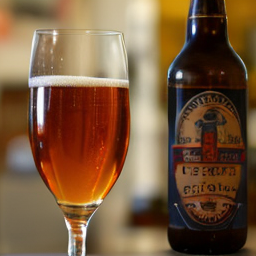}   &
    \includegraphics[width=.24\textwidth]{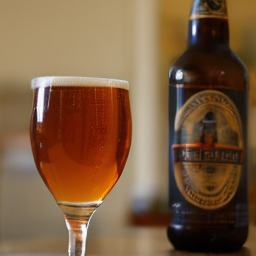}   &
    \includegraphics[width=.24\textwidth]{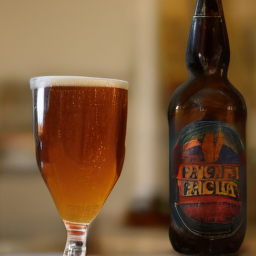}  \\[2pt]

    \includegraphics[width=.24\textwidth]{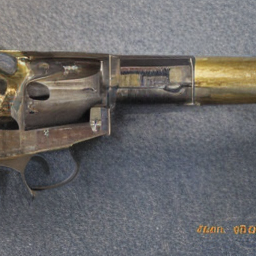} &
    \includegraphics[width=.24\textwidth]{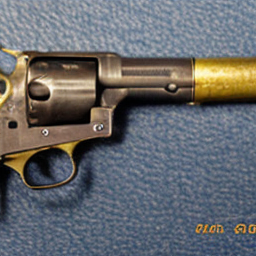}   &
    \includegraphics[width=.24\textwidth]{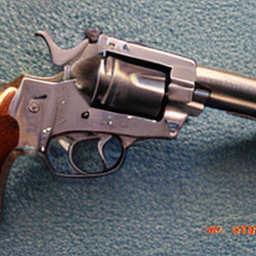}   &
    \includegraphics[width=.24\textwidth]{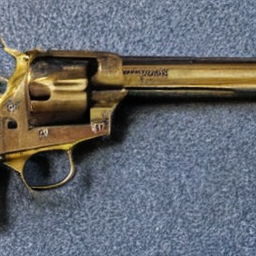}
  \end{tabular}
  \caption{Qualitative samples using a guidance scale of $\omega = 2.5$ across different methods: Unguided, Sparse Guidance (SG), AutoGuidance (AG), and Classifier-Free Guidance (CFG).}
  \label{fig:samples2}
\end{figure*}

\begin{figure*}[h]
  \centering
  \setlength{\tabcolsep}{1pt}
  \begin{tabular}{cccc}
    \textbf{Unguided} & \textbf{SG} & \textbf{AG} & \textbf{CFG} \\[2pt]
    \includegraphics[width=.24\textwidth]{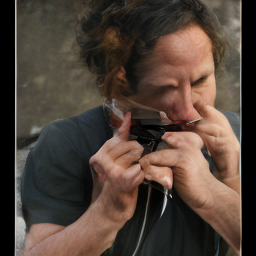} &
    \includegraphics[width=.24\textwidth]{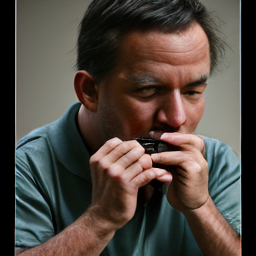}   &
    \includegraphics[width=.24\textwidth]{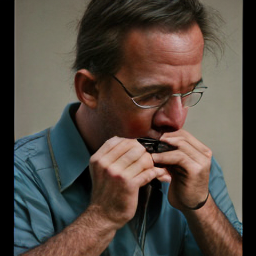}   &
    \includegraphics[width=.24\textwidth]{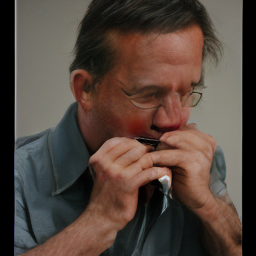}  \\[2pt]

    \includegraphics[width=.24\textwidth]{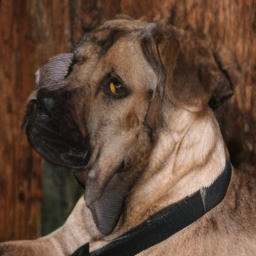} &
    \includegraphics[width=.24\textwidth]{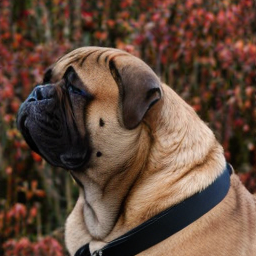}   &
    \includegraphics[width=.24\textwidth]{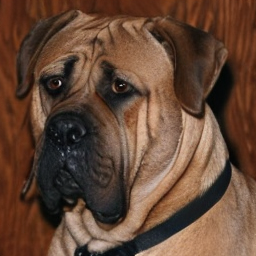}   &
    \includegraphics[width=.24\textwidth]{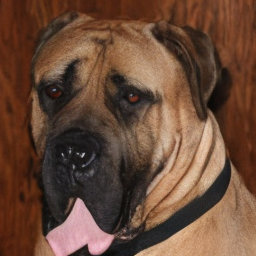}  \\[2pt]

    \includegraphics[width=.24\textwidth]{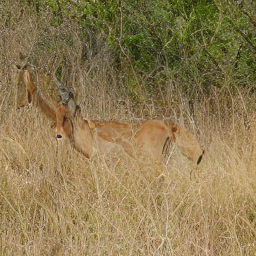} &
    \includegraphics[width=.24\textwidth]{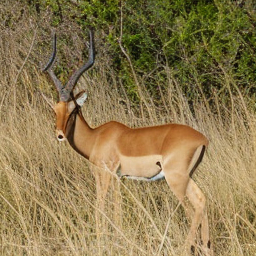}   &
    \includegraphics[width=.24\textwidth]{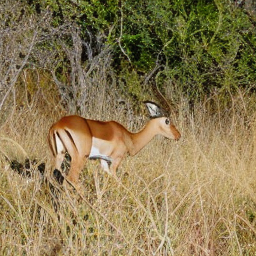}   &
    \includegraphics[width=.24\textwidth]{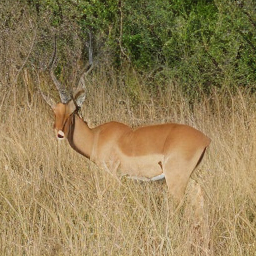}  \\[2pt]

    \includegraphics[width=.24\textwidth]{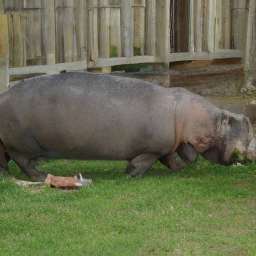} &
    \includegraphics[width=.24\textwidth]{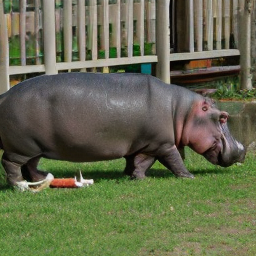}   &
    \includegraphics[width=.24\textwidth]{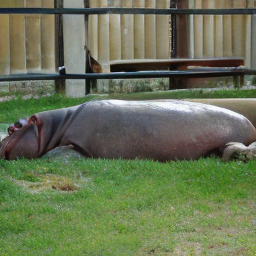}   &
    \includegraphics[width=.24\textwidth]{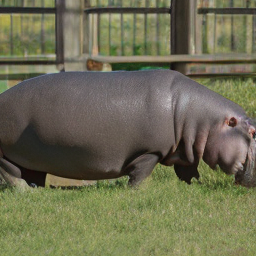}
  \end{tabular}
  \caption{Qualitative samples using a guidance scale of $\omega = 2.5$ across different methods: Unguided, Sparse Guidance (SG), AutoGuidance (AG), and Classifier-Free Guidance (CFG).}
  \label{fig:samples1}
\end{figure*}

\begin{figure*}[ht]
    \centering
    \rowpair{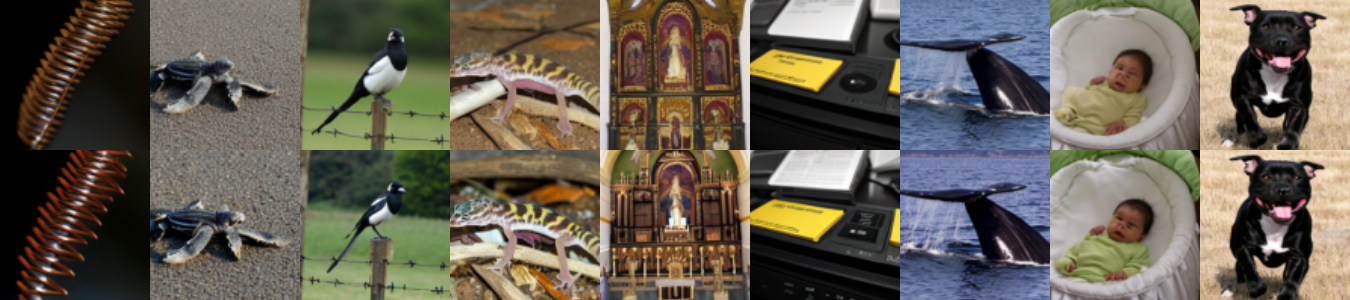}\par\vspace{10mm}
    \rowpair{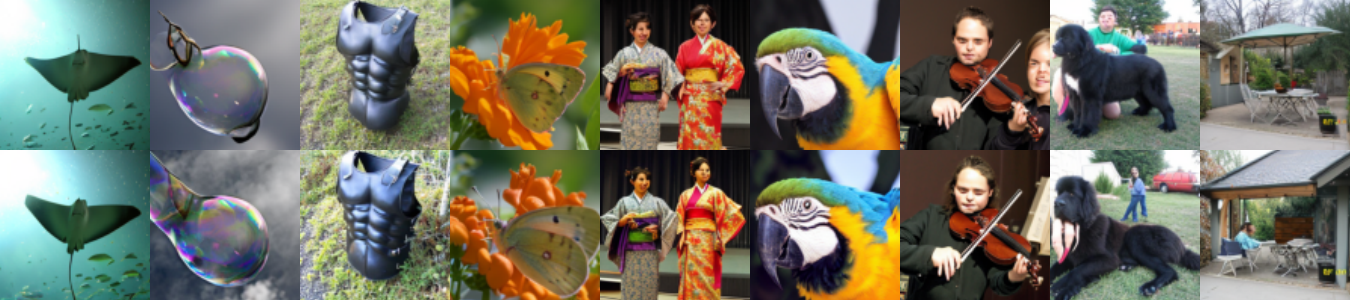}\par\vspace{10mm}
    \rowpair{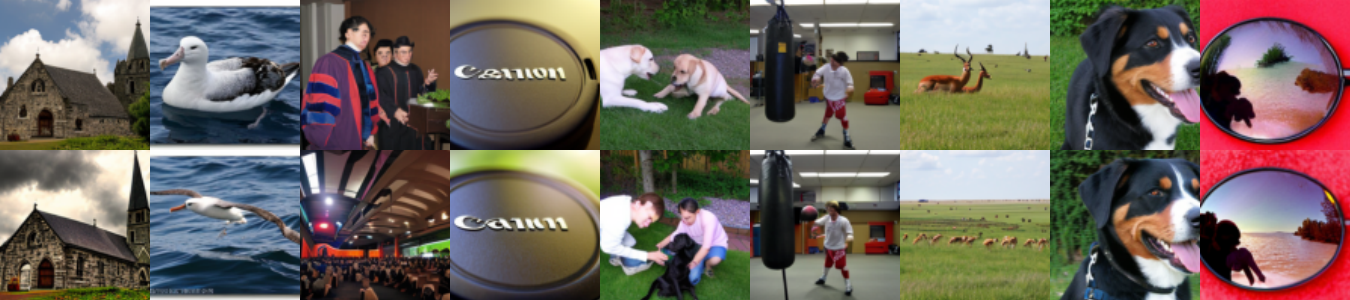}\par\vspace{10mm}
    \rowpair{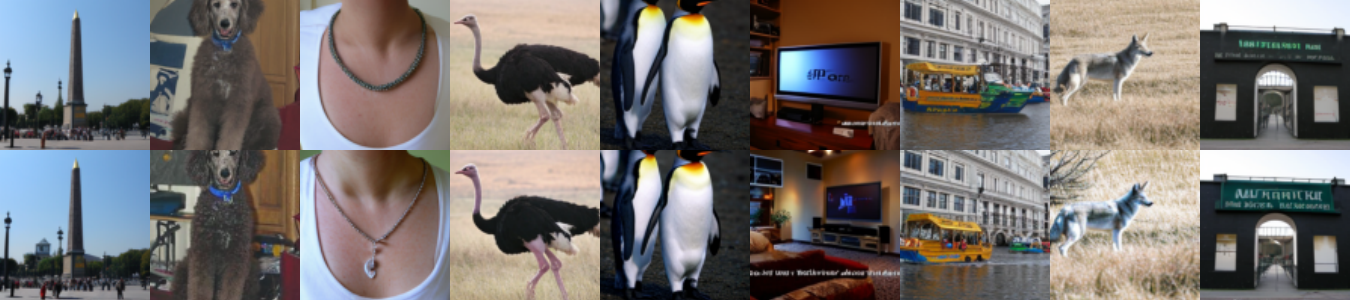}
    \vspace{1em}
    \caption{\textbf{Uncurated samples} of $\text{SG}_{\text{FLOPS}}$ (top) and $\text{SG}_{\text{FID}}$ (bottom) using $\omega=2.5$.}
    \label{fig:rows1}
\end{figure*}

\begin{figure*}[ht]
    \centering
    \rowpair{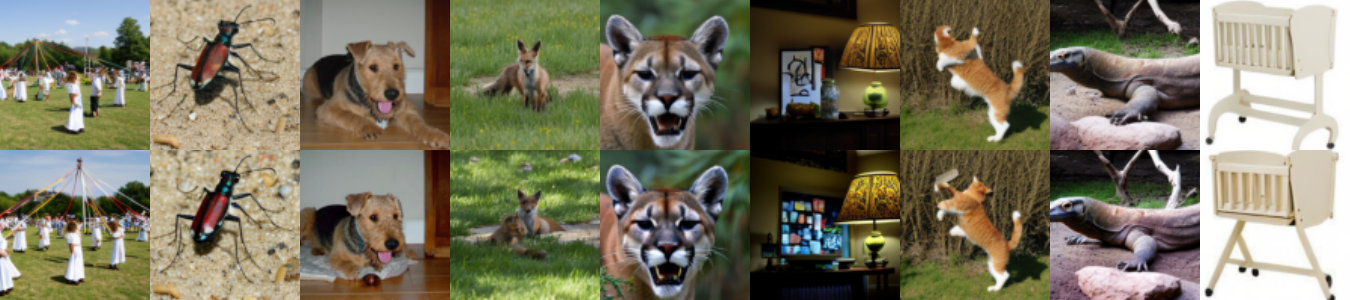}\par\vspace{10mm}
    \rowpair{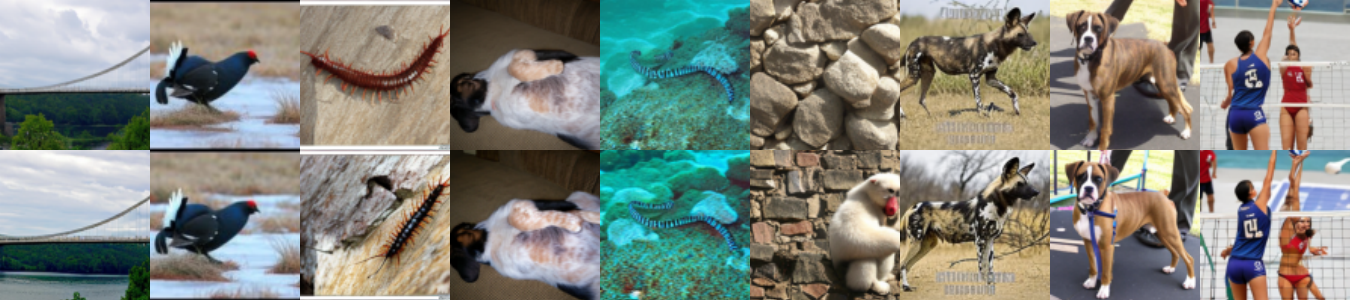}\par\vspace{10mm}
    \rowpair{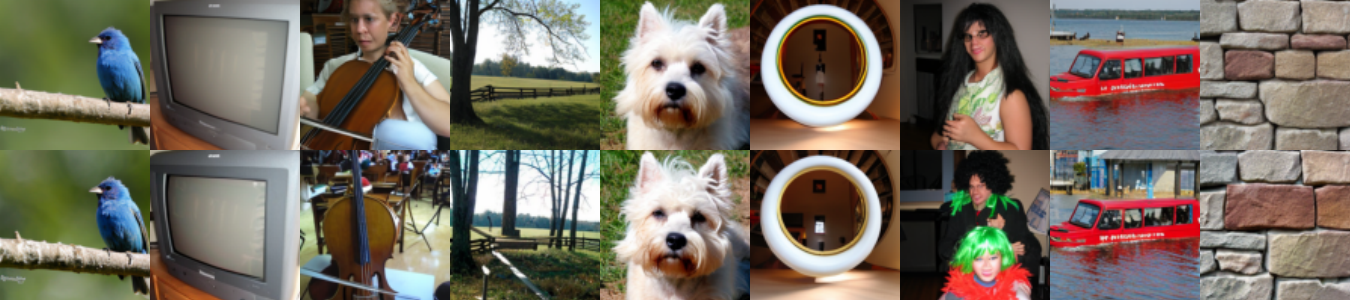}\par\vspace{10mm}
    \rowpair{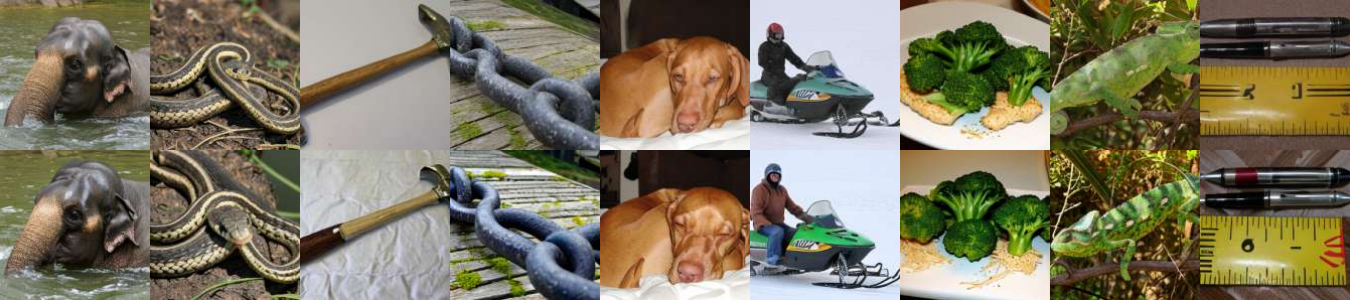}
    \vspace{1em}
    \caption{\textbf{Uncurated samples} of $\text{SG}_{\text{FLOPS}}$ (top) and $\text{SG}_{\text{FID}}$ (bottom) using $\omega=2.5$.}
    \label{fig:rows2}
\end{figure*}

\begin{figure*}[ht]
    \centering
    \rowpair{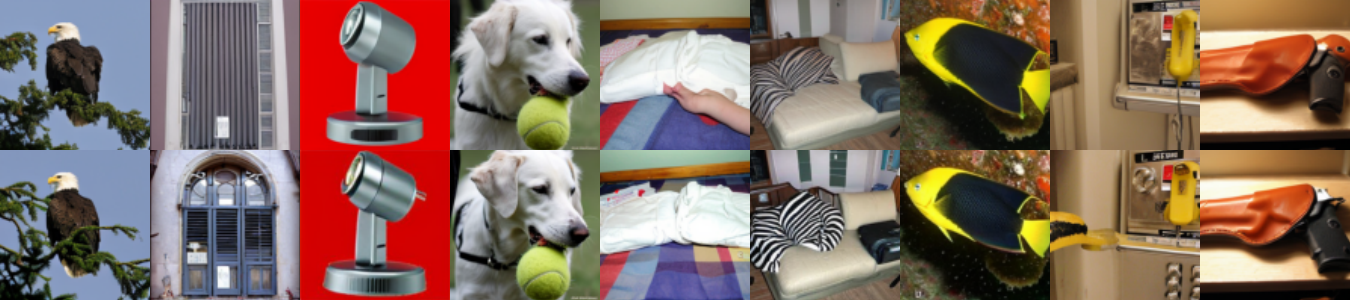}\par\vspace{10mm}
    \rowpair{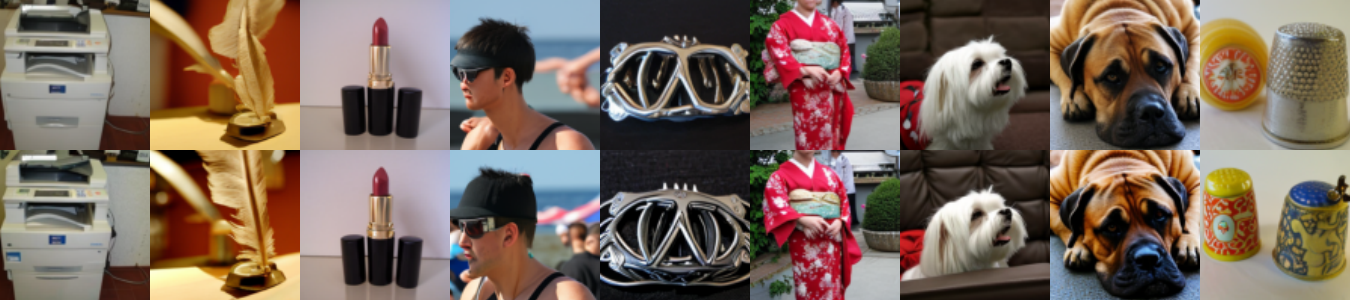}\par\vspace{10mm}
    \rowpair{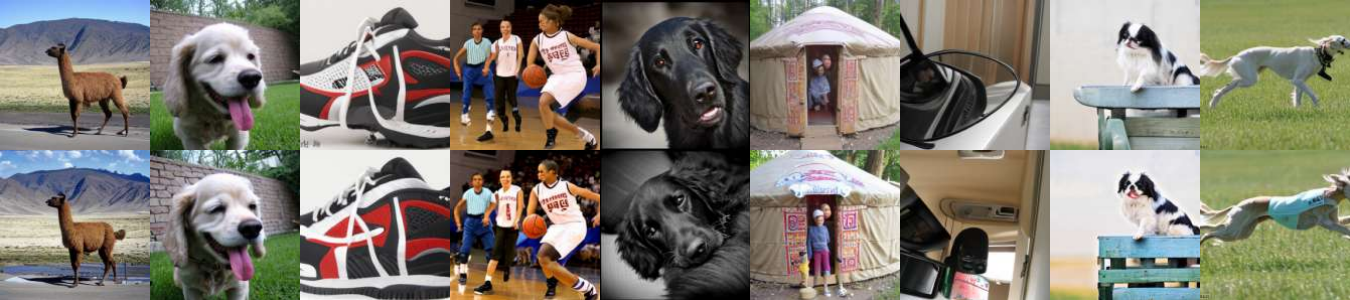}\par\vspace{10mm}
    \rowpair{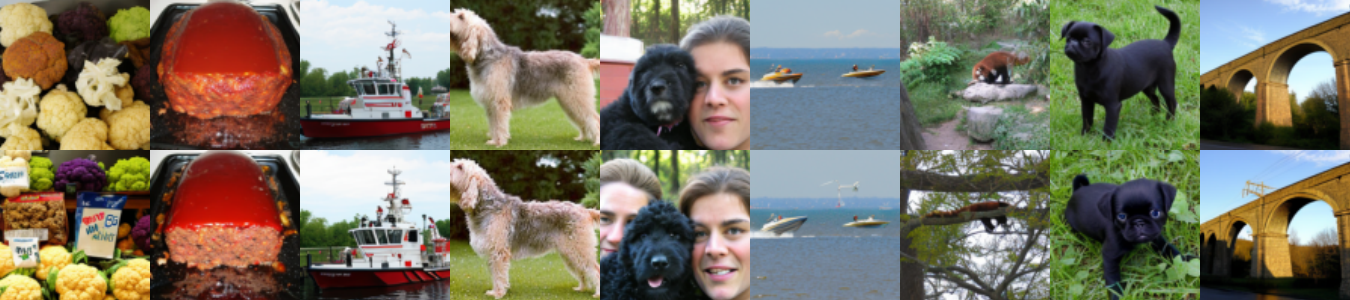}
    \vspace{1em}
    \caption{\textbf{Uncurated samples} of $\text{SG}_{\text{FLOPS}}$ (top) and $\text{SG}_{\text{FID}}$ (bottom) using $\omega=2.5$.}
    \label{fig:rows3}
\end{figure*}

Subsequently, \Cref{fig:samples2} and \Cref{fig:samples1} display ImageNet-256 results, contrasting unguided predictions, AutoGuidance (AG), CFG, and our SG method. Finally, \Cref{fig:rows1}, \Cref{fig:rows2}, and \Cref{fig:rows3} offer uncurated qualitative comparisons between $\text{SG}_{\text{FID}}$ and $\text{SG}_{\text{FLOPS}}$ to illustrate their respective visual characteristics.

\begin{figure*}
    \centering
    \begin{tikzpicture}
        \def\leftmargin{44pt}   
        \def\topoffset{5pt}
        \def\leftoffset{5pt}

        \node[inner sep=0, anchor=north west] (img) at (\leftmargin,0) {%
            \includegraphics[width=\dimexpr\textwidth-\leftmargin\relax]{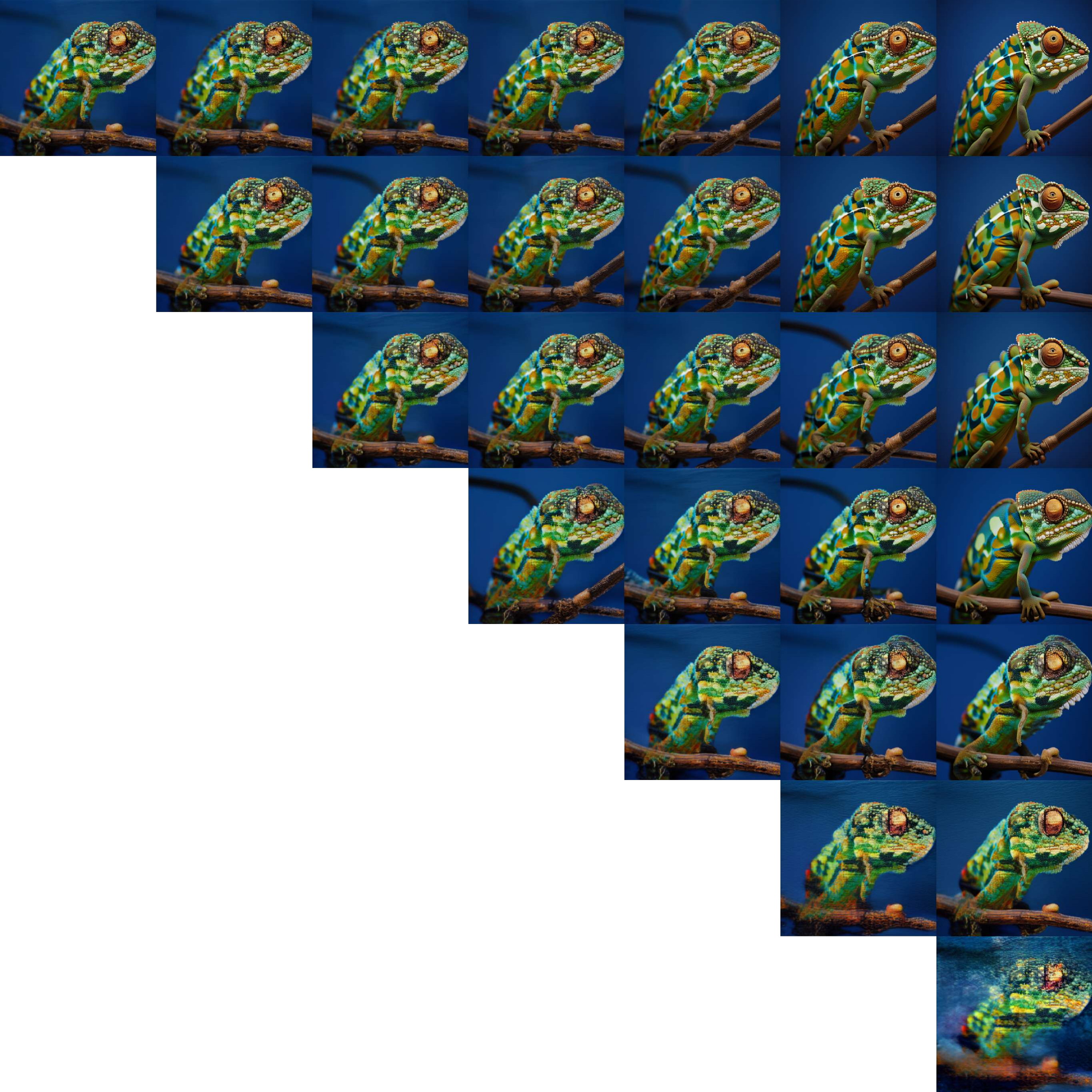}%
        };

        \foreach[count=\i] \val in {0.0,0.1,0.2,0.4,0.6,0.8,0.9} {
            \pgfmathsetmacro{\xpos}{(\i-0.5)/7}
            \node[font=\footnotesize, anchor=south]
                at ($(img.north west)!\xpos!(img.north east) + (0,\topoffset)$) {\val};
        }

        \foreach[count=\i] \val in {0.0,0.1,0.2,0.4,0.6,0.8,0.9} {
            \pgfmathsetmacro{\ypos}{(\i-0.5)/7}
            \node[font=\footnotesize, anchor=east]
                at ($(img.north west)!\ypos!(img.south west) + (-\leftoffset,0)$) {\val};
        }

        \node[font=\large\bfseries, anchor=south]
            at ($(img.north west)!0.5!(img.north east) + (0,18pt)$)
            {$\gamma_{\mathrm{weak}}$};

        \node[font=\large\bfseries, rotate=90, anchor=south]
            at ($(img.north west)!0.5!(img.south west) + (-26pt,0)$)
            {$\gamma_{\mathrm{strong}}$};
    \end{tikzpicture}
    \caption{\textbf{Effect of Sparse Guidance on image quality.} Lower values for $\gamma_{\mathrm{strong}}$ and higher ones for $\gamma_{\mathrm{weak}}$ lead to best quality. \textit{Prompt: The image showcases a vibrant chameleon perched on a slender branch against a striking, deep blue background. The chameleon is the main focus, presented in a close-up, side-view shot that emphasizes its fascinating textures and colors. Its skin is a tapestry of greens, blues, and browns, creating a mottled pattern of scales. Patches of white punctuate the green, and larger blotches of a reddish-brown add depth to its coloration. The head features a gradient of yellow and orange, drawing attention to its unique eye. The eye itself is a complex mix of orange and black. The texture of the chameleon's skin is visibly bumpy and scaled. Its curled toes are gripping the branch. The overall impression is one of natural beauty and intricate detail. The sharp focus and simple background allow for a detailed examination of the chameleon's unique features.}}
    \label{fig:grid1}
\end{figure*}

\begin{figure*}
    \centering
    \begin{tikzpicture}
        \def\leftmargin{44pt}   
        \def\topoffset{5pt}
        \def\leftoffset{5pt}

        \node[inner sep=0, anchor=north west] (img) at (\leftmargin,0) {%
            \includegraphics[width=\dimexpr\textwidth-\leftmargin\relax]{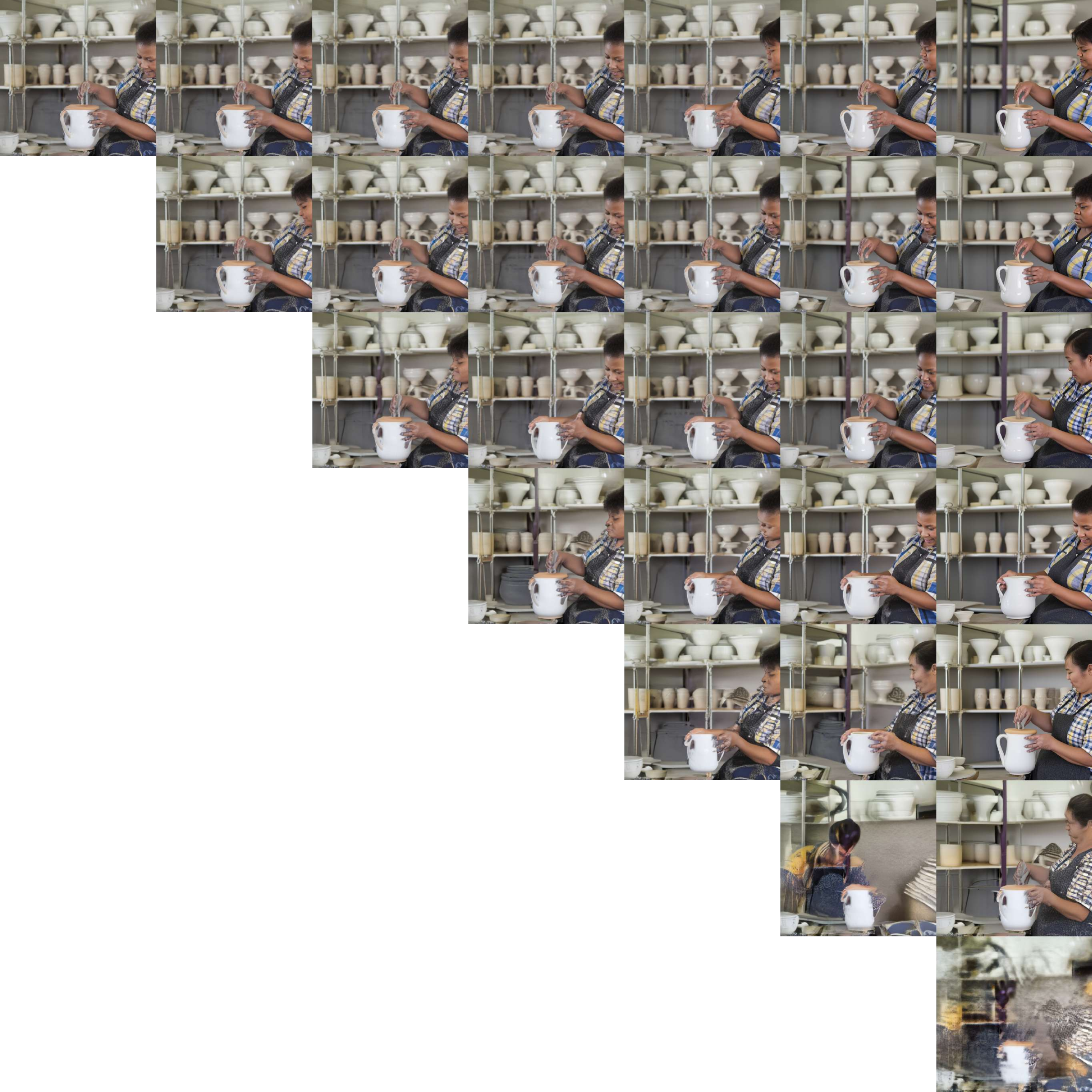}%
        };

        \foreach[count=\i] \val in {0.0,0.1,0.2,0.4,0.6,0.8,0.9} {
            \pgfmathsetmacro{\xpos}{(\i-0.5)/7}
            \node[font=\footnotesize, anchor=south]
                at ($(img.north west)!\xpos!(img.north east) + (0,\topoffset)$) {\val};
        }

        \foreach[count=\i] \val in {0.0,0.1,0.2,0.4,0.6,0.8,0.9} {
            \pgfmathsetmacro{\ypos}{(\i-0.5)/7}
            \node[font=\footnotesize, anchor=east]
                at ($(img.north west)!\ypos!(img.south west) + (-\leftoffset,0)$) {\val};
        }

        \node[font=\large\bfseries, anchor=south]
            at ($(img.north west)!0.5!(img.north east) + (0,18pt)$)
            {$\gamma_{\mathrm{weak}}$};

        \node[font=\large\bfseries, rotate=90, anchor=south]
            at ($(img.north west)!0.5!(img.south west) + (-26pt,0)$)
            {$\gamma_{\mathrm{strong}}$};
    \end{tikzpicture}
    \caption{\textbf{Effect of Sparse Guidance on image quality.} Lower values for $\gamma_{\mathrm{strong}}$ and higher ones for $\gamma_{\mathrm{weak}}$ lead to best quality. \textit{Prompt: The image shows a person engaged in the process of pottery making, specifically sanding a white ceramic pitcher. The individual, wearing a blue and yellow plaid shirt and a speckled black apron, carefully holds the pitcher with one hand while using a piece of sandpaper to smooth its surface with the other. In the background, a metal shelving unit filled with stacks of white ceramic bowls and cups indicates a workshop or studio setting. More unglazed ceramic pieces, including cups and plates, sit on a surface in the foreground, suggesting a production line or batch of pottery in progress. The lighting is soft and natural, highlighting the details of the ceramic surfaces and the craftsman's hands. The overall composition focuses on the tactile and meticulous nature of pottery making.}}
    \label{fig:grid2}
\end{figure*}

\end{document}